\documentclass[11pt]{article}
\usepackage[preprint]{acl}
\usepackage{times}
\usepackage{latexsym}
\usepackage[T1]{fontenc}
\usepackage[utf8]{inputenc}
\usepackage{microtype}
\usepackage{inconsolata}
\usepackage{graphicx}
\usepackage{amsmath, amssymb, amsthm}
\usepackage{mathtools}
\usepackage{bm}
\usepackage{booktabs}
\usepackage{enumitem}
\usepackage{hyperref}
\usepackage{cleveref}
\usepackage{natbib}
\usepackage{titletoc}
\usepackage{subcaption}
\usepackage{multirow}
\usepackage{array}
\usepackage{tikz}
\usepackage{xcolor}
\usepackage{eso-pic,xcolor}

\usetikzlibrary{arrows.meta, shapes, positioning, fit, backgrounds}

\hypersetup{colorlinks=true, linkcolor=blue, citecolor=blue, urlcolor=blue}

\definecolor{nohumancolor}{RGB}{180,180,180}
\definecolor{neutralcolor}{RGB}{66,133,244}
\definecolor{biasedcolor}{RGB}{219,68,55}
\definecolor{presencecolor}{RGB}{66,133,244}
\definecolor{contentcolor}{RGB}{219,68,55}
\definecolor{totalcolor}{RGB}{15,157,88}
\definecolor{boxbg}{RGB}{248,249,250}
\definecolor{arrowgray}{RGB}{100,100,100}

\title{Conditional Cognitive Biases in LLMs:\\
How Biased User Turns Modulate In-Context Reasoning}

\author{
Sachini Weerasekara,
Sagar Kamarthi, \and
Jacqueline Isaacs \\[0.5em]
Northeastern University \\[0.5em]
\texttt{\{s.weerasekara, sagar, isaacs\}@northeastern.edu}
}

\begin{document}
\maketitle

\begin{abstract}
We present an evaluation of cognitive bias expression in state-of-the-art instruction-tuned LLMs under realistic multi-turn interaction settings. Our work introduces a novel three-condition experimental framework that disentangles the effect of exposure to a biased user turn from the effect of the turn’s semantic content, alongside a benchmark of 24{,}300 jury-validated user prompts spanning all 81 cells of a $9 \times 9$ target--human bias interaction matrix. Across eight frontier LLMs, we find that biased conversational context systematically increases bias expression relative to zero-shot baselines in 6 of 8 models. We identify two competing behavioral dynamics underlying this effect: conversational exposure to biased reasoning generally amplifies downstream bias tendencies, while explicitly stated bias cues often trigger alignment-related suppression behaviors that reduce overt bias expression. We release our framework, codebase, and dataset to support future research on context-conditioned cognitive biases and behavioral adaptation in LLMs.
\end{abstract}

\section{Introduction}

Instruction-tuned large language models are increasingly deployed as conversational decision-support tools in high-stakes settings including legal reasoning, medical triage, and financial advisory contexts \citep{bommasani2021opportunities,achiam2023gpt,dubey2024llama}. In these deployments, users routinely exhibit systematic cognitive biases, framing, anchoring, optimism bias, that shape how requests are formulated and how information is presented to the model \citep{kahneman2011thinking,tversky1974judgment,tversky1981framing}. Whether and how these biased user turns propagate into downstream model behavior remains an open question.

A substantial body of work has characterized cognitive bias in LLMs under zero-shot evaluation \citep{malberg2025comprehensive,jones2022capturing,gallegos2024bias,binz2023using,hagendorff2023human}, establishing susceptibility to framing \citep{tversky1981framing}, anchoring \citep{strack1997explaining,epley2001putting}, in-group preference \citep{tajfel1979integrative}, and confirmation bias \citep{nickerson1998confirmation, weerasekara2026prototype, weerasekara2025improvements}. These benchmarks have been pivotal in demonstrating that alignment training does not eliminate bias \citep{perez2022discovering,ouyang2022training,santurkar2023whose}. Yet every benchmark result is measured from an isolated, carefully controlled prompt, treating LLM bias as a context-independent property. This does not reflect deployment conditions, where model responses to decision-critical prompts are always conditioned on a preceding conversational turn.

A biased user turn confounds two causal pathways: the \emph{presence} of any user turn changes the input from zero-shot to dialogue, activating instruction-tuning conditioning absent at zero-shot time; the \emph{content} of a biased turn introduces a second signal that interacts with the model's alignment recipe to amplify or suppress the target bias. Without separating these factors, the mechanisms driving changes in LLM bias expression remain opaque.

To isolate these pathways, we introduce a three-condition design with \textbf{no user turn} ($m_\varnothing$), \textbf{neutral user turn} ($m_n$), and \textbf{biased user turn} ($m_b$). This yields the primary estimand $\Delta_b = |m_b| - |m_\varnothing|$ and its decomposition $\Delta_b = \Delta_n + \delta$, where the presence effect $\Delta_n = |m_n| - |m_\varnothing|$ isolates format shift and the content effect $\delta = |m_b| - |m_n|$ isolates bias-specific modulation. We construct a stimulus bank of 24,300 jury-validated turns covering all 81 cells of a $9\!\times\!9$ matrix of target LLM biases $\beta$ against human-turn biases $\gamma$, evaluated across 8 instruction-tuned LLMs spanning diverse parameter scales and alignment recipes. A three-model LLM-as-judge jury (GPT-4o-Mini, Claude-3.5-Haiku, Gemini) enforces four quality criteria before any stimulus enters the bank, decoupling stimulus generation from evaluation. Causal effects are identified via difference-in-differences \citep{angrist2009mostly}, synthetic control \citep{abadie2010synthetic}, and propensity score matching \citep{rosenbaum1983central}.

We ask: \textbf{To what extent does the cognitive bias content of a user turn modulate LLM bias expression beyond the format shift induced by any user turn, and does a human's specific bias type selectively induce the matching bias in the model?}

We find that biased user turns elevate LLM bias above zero-shot levels ($\bar{\Delta}_b > 0$) in 6 of 8 models (range $+0.022$--$+0.051$; $p{<}0.001$), with a significant presence effect across all 8 models (positive in 7, negative in Claude-3.5-Haiku) partially counteracted by predominantly negative content effects (suppression in 6 of 8 models), with deliberative models (DeepSeek-V3, Claude-3.5-Haiku) reversing the content direction. The $9\!\times\!9$ coupling matrix reveals strong row structure but no diagonal advantage, inducibility is governed by the target bias, not the human-to-LLM bias pairing, and Planning Fallacy is the only target bias causally confirmed as universally inducible across all 8 models.

\section{Related Work}

\paragraph{Cognitive bias benchmarking in LLMs.}
Prior work catalogs LLM cognitive biases, framing \citep{tversky1981framing}, anchoring \citep{strack1997explaining,epley2001putting}, availability \citep{tversky1973availability}, in-group bias \citep{tajfel1979integrative}, confirmation bias \citep{nickerson1998confirmation}, under zero-shot prompting \citep{malberg2025comprehensive,jones2022capturing,gallegos2024bias}, fixing evaluation to a single isolated prompt \citep{liang2022holistic} and treating bias as context-independent. We move beyond this paradigm by measuring how a single in-context user turn modulates downstream bias expression.

\paragraph{In-context learning and sycophancy.}
LLMs update their output distribution from context-prepended demonstrations without gradient updates \citep{brown2020language,dong2023survey}; format and input distribution matter more than label semantics \citep{min2022rethinking,zhao2021calibrate}, and prompt ordering alone shifts accuracy substantially \citep{lu2022fantastically}. Instruction-tuned models exhibit sycophancy, changing answers when users express disagreement \citep{sharma2023towards}, and chain-of-thought reasoning can be post-hoc rationalized by biased context \citep{turpin2023language}. LLMs also reproduce survey response biases characteristic of human respondents \citep{tjuatja2024llms, weerasekara2024reinforcement, weerasekara2022trends}. Model-written evaluations confirm that behavioral biases persist under alignment \citep{perez2022discovering,santurkar2023whose, weerasekara2025cellclique}. Neither line places the \emph{cognitive bias type} of the user turn under systematic control at the scale we study.

\paragraph{LLMs as evaluation data generators and causal methods.}
Persona-injection prompting generates human-like cognitive profiles for evaluation \citep{argyle2023out}; we add a three-model LLM-as-judge jury \citep{zheng2023judging} to enforce stimulus quality. We apply difference-in-differences \citep{angrist2009mostly}, synthetic control \citep{abadie2010synthetic}, and propensity score matching \citep{rosenbaum1983central} to provide per-bias causal evidence beyond aggregate comparisons.

\section{Methodology}

\subsection{Formal Setup}

Let $\mathcal{B}$ denote the nine target cognitive biases. For scenario $x \in \mathcal{X}$, the biased user turn is an intervention operator $\mathcal{I}_\gamma^x : x \mapsto c_\gamma$ that prepends a user turn embedding bias $\gamma$ to the model's context. The primary estimand is:
\[
\Pr\!\bigl(Y_x^\beta = 1 \mid c_\gamma\bigr) - \Pr\!\bigl(Y_x^\beta = 1 \mid c_\varnothing\bigr),
\]
the shift in bias $\beta$ expression probability relative to the zero-shot baseline $c_\varnothing$.

\subsection{Bias Strength Metric}

We adopt the normalized ratio metric of \citet{malberg2025comprehensive}. Each scenario $x$ is evaluated by eliciting two numerical decisions from the LLM: $a_1$ (response to the \emph{control} template, neutral framing) and $a_2$ (response to the \emph{treatment} template, bias-inducing framing). Reference anchors $y_1, y_2 \in \mathbb{R}$ from per-scenario \texttt{metric\_params} allow comparison across heterogeneous scales. Define $\Delta_i = a_i - y_i$. The bias metric is:
\begin{equation}
m(a_{1,2}, y_{1,2}, k)
= \frac{k \cdot \bigl(|\Delta_1| - |\Delta_2|\bigr)}{\max(|\Delta_1|, |\Delta_2|)}, \quad m \in [-1, 1],
\label{eq:metric}
\end{equation}
where $k \in \{-1, +1\}$ corrects for scenario-level direction. Values near $+1$ indicate strong bias; near $0$, negligible bias. Both $a_1$ and $a_2$ are elicited in every condition, so $m$ measures the \emph{differential sensitivity} to scenario framing within each conversational context, not a raw decision value.

\subsection{Three-Condition Evaluation Design}

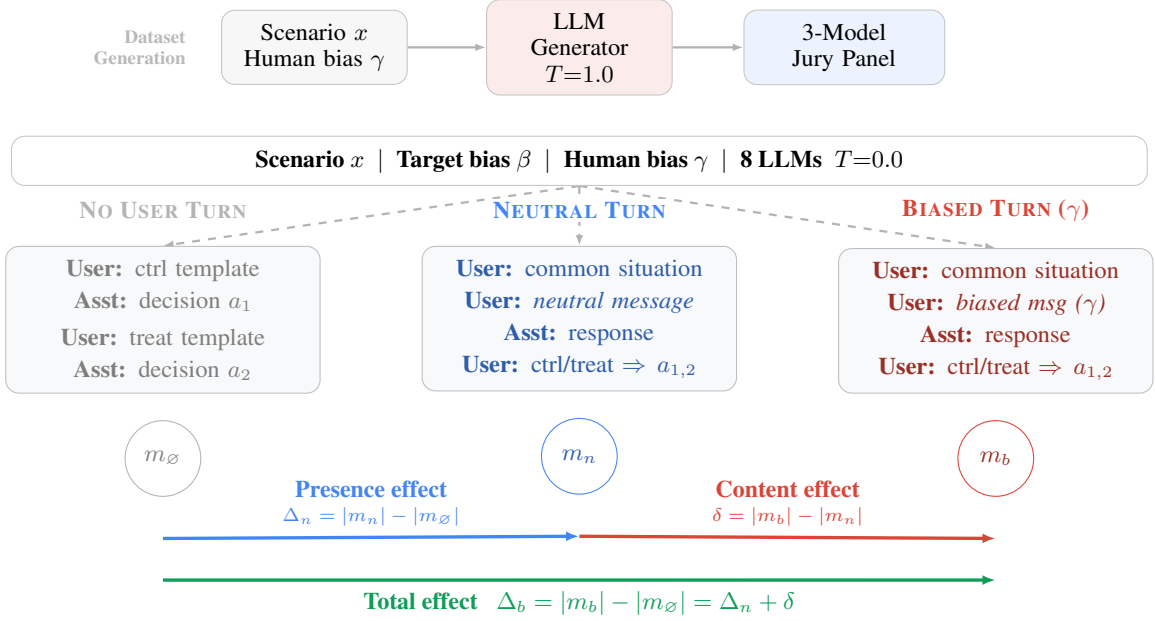
\begin{figure*}[t]
\centering
\begin{tikzpicture}[
  font=\small,
  every node/.style={align=center},
  convbox/.style={
    draw=gray!50, rounded corners=5pt, fill=boxbg,
    minimum width=4.0cm, inner sep=5pt, text width=3.8cm,
    font=\footnotesize
  },
  scorebox/.style={
    draw, circle, fill=white,
    minimum size=1.0cm, font=\footnotesize\bfseries,
    inner sep=2pt
  },
  effectlabel/.style={font=\footnotesize\bfseries, midway},
  colhead/.style={font=\small\bfseries, fill=white, inner sep=1pt},
  stepbox/.style={
    draw=gray!45, rounded corners=5pt,
    minimum width=2.2cm, minimum height=0.82cm,
    inner sep=5pt, font=\footnotesize, text width=2.1cm
  },
  arr/.style={-{Latex[length=4pt]}, gray!60, line width=0.85pt},
  effarr/.style={-{Latex[length=5pt]}, line width=1.25pt},
]


\node[font=\scriptsize\bfseries, text=gray!55, align=right, text width=1.6cm]
  at (-6.0, 2.1) {Dataset\\Generation};

\node[stepbox, fill=gray!6]           (inp)  at (-3.5, 2.1) {Scenario $x$\\Human bias $\gamma$};
\node[stepbox, fill=biasedcolor!10]   (gen)  at ( 0.0, 2.1) {LLM\\Generator\\$T{=}1.0$};
\node[stepbox, fill=neutralcolor!10, text width=2.3cm, minimum width=2.5cm]
                                      (jury) at ( 3.5, 2.1) {3-Model\\Jury Panel};

\draw[arr] (inp) -- (gen);
\draw[arr] (gen) -- (jury);


\node[draw=gray!45, rounded corners=5pt, fill=white, minimum width=15.0cm,
      minimum height=0.65cm, font=\footnotesize\bfseries] (scenario) at (0, 0.6)
  {Scenario $x$ \;$|$\; Target bias $\beta$ \;$|$\; Human bias $\gamma$ \;$|$\; 8 LLMs \;$T{=}0.0$};


\node[colhead, text=nohumancolor] (hd1) at (-5.5, -0.05) {\textsc{No User Turn}};
\node[colhead, text=neutralcolor] (hd2) at (  0.0, -0.05) {\textsc{Neutral Turn}};
\node[colhead, text=biasedcolor]  (hd3) at (  5.5, -0.05) {\textsc{Biased Turn ($\gamma$)}};


\node[convbox, below=0.32cm of hd1, text=nohumancolor!70!black] (nh_box) {
  \textbf{User:} ctrl template\\[2pt]
  \textbf{Asst:} decision $a_1$\\[4pt]
  \textbf{User:} treat template\\[2pt]
  \textbf{Asst:} decision $a_2$
};
\node[convbox, below=0.32cm of hd2, text=neutralcolor!70!black] (n_box) {
  \textbf{User:} common situation\\[2pt]
  \textbf{User:} \textit{neutral message}\\[2pt]
  \textbf{Asst:} response\\[2pt]
  \textbf{User:} ctrl/treat $\Rightarrow a_{1,2}$
};
\node[convbox, below=0.32cm of hd3, text=biasedcolor!70!black] (b_box) {
  \textbf{User:} common situation\\[2pt]
  \textbf{User:} \textit{biased msg ($\gamma$)}\\[2pt]
  \textbf{Asst:} response\\[2pt]
  \textbf{User:} ctrl/treat $\Rightarrow a_{1,2}$
};


\node[scorebox, below=0.35cm of nh_box, draw=nohumancolor, text=nohumancolor!70!black]
  (s0) {$m_\varnothing$};
\node[scorebox, below=0.35cm of n_box,  draw=neutralcolor, text=neutralcolor!70!black]
  (sn) {$m_n$};
\node[scorebox, below=0.35cm of b_box,  draw=biasedcolor,  text=biasedcolor!70!black]
  (sb) {$m_b$};


\begin{pgfonlayer}{background}
  \foreach \nd in {nh_box, n_box, b_box}{
    \draw[arr, dashed] (scenario.south) -- (\nd.north);
  }
\end{pgfonlayer}


\draw[effarr, presencecolor]
  ([yshift=-0.55cm]s0.south) --
  node[effectlabel, above, text=presencecolor, align=center] {
    Presence effect\\[-1pt]
    {\normalfont\scriptsize$\Delta_n = |m_n| - |m_\varnothing|$}
  }
  ([yshift=-0.55cm]sn.south);

\draw[effarr, contentcolor]
  ([yshift=-0.55cm]sn.south) --
  node[effectlabel, above, text=contentcolor, align=center] {
    Content effect\\[-1pt]
    {\normalfont\scriptsize$\delta = |m_b| - |m_n|$}
  }
  ([yshift=-0.55cm]sb.south);

\draw[effarr, totalcolor]
  ([yshift=-1.10cm]s0.south) --
  node[effectlabel, below, text=totalcolor]
    {Total effect \; $\Delta_b = |m_b| - |m_\varnothing| = \Delta_n + \delta$}
  ([yshift=-1.10cm]sb.south);

\end{tikzpicture}
\caption{Experimental pipeline. For each (scenario $x$, target bias $\beta$, human bias $\gamma$) triple, the LLM is evaluated under three conditions. The bias metric $m$ (Eq.~\ref{eq:metric}) pairs the control decision $a_1$ with the treatment decision $a_2$ within each condition. Two effects are isolated: the \textbf{presence effect} ($|m_n| - |m_\varnothing|$) attributable to conversational format alone, and the \textbf{content effect} ($|m_b| - |m_n|$) attributable to the bias-specific content of the user turn. The total effect ($|m_b| - |m_\varnothing|$) combines both. All effects are differences of absolute bias magnitudes.}
\label{fig:pipeline}
\end{figure*}

A biased user turn confounds two causal pathways: format (any user turn vs.\ none) and content (biased vs.\ neutral). The neutral condition holds format constant while varying content, making both pathways separately estimable; without it, $|m_b| - |m_\varnothing|$ cannot be attributed to bias content alone.

For each (scenario, human-bias) pair $(x, \gamma)$, the instruction-tuned LLM is therefore evaluated under three conditions (Figure~\ref{fig:pipeline}):

\medskip
\noindent\textbf{Zero-shot / no user turn} ($m_\varnothing$): Control and treatment templates are presented as independent single-turn prompts with no prior context. This replicates the standard evaluation design of \citet{malberg2025comprehensive} and establishes the model's intrinsic zero-shot bias baseline.

\medskip
\noindent\textbf{Neutral user turn} ($m_n$): A multi-turn chat context is constructed in which a bias-free user turn precedes the decision prompt. This isolates the effect of user-turn presence in the context window from any bias-specific semantic content.

\medskip
\noindent\textbf{Biased user turn} ($m_b$): Same multi-turn chat structure, but the in-context user turn expresses bias $\gamma$ (LLM-as-judge validated). Following the standard chat template format:
\begin{enumerate}[leftmargin=*, topsep=1pt, itemsep=0pt, label=\arabic*.]
\item \textbf{User}: Common situation prefix (shared context across control and treatment)
\item \textbf{User}: User turn (neutral or biased, generated via persona-injection prompting)
\item \textbf{Assistant}: LLM's intermediate completion (acknowledgment)
\item \textbf{User}: Template-diverging text + decision prompt + answer options
\item \textbf{Assistant}: LLM's final decision completion ($a_1$ or $a_2$)
\end{enumerate}
Steps 1--5 are run twice per condition, once with the control template's diverging text (eliciting $a_1$) and once with the treatment's (eliciting $a_2$). The resulting bias score $m$ reflects in-context framing sensitivity under that conversational prefix.

\subsection{Effect Decomposition}

The three conditions yield two orthogonal contrasts and one composite, all defined on absolute bias magnitudes:
\begin{align}
\text{Presence effect:} &\quad \Delta_n = |m_n| - |m_\varnothing| \label{eq:presence}\\
\text{Content effect:}  &\quad \delta   = |m_b| - |m_n| \label{eq:content}\\
\text{Total effect:}    &\quad \Delta_b = |m_b| - |m_\varnothing| = \Delta_n + \delta \label{eq:total}
\end{align}
$\Delta_n$ isolates format (user turn vs.\ none, content neutral); $\delta$ isolates bias content (biased vs.\ neutral turn, format constant); $\Delta_b = \Delta_n + \delta$ is what a two-condition study collapses. Absolute differences ensure $\delta > 0$ ($\delta < 0$) unambiguously means increased (decreased) bias magnitude, avoiding sign artefacts from shifts between anti-biased and neutral completions.

\section{Experimental Setup}

\subsection{Bias Selection}

We evaluate nine cognitive biases drawn from the 30-bias benchmark of \citet{malberg2025comprehensive}: Anchoring, Availability Heuristic, Bandwagon Effect, Confirmation Bias, Framing Effect, In-Group Bias, Loss Aversion, Planning Fallacy, and Status Quo Bias. Selection criteria are: (1) reliable single-turn signal in \citet{malberg2025comprehensive}; and (2) established mechanistic links in the human cognition literature for theoretically motivated cross-bias pairs, framing--loss aversion \citep{kahneman2013prospect,tversky1981framing}, loss aversion--status quo \citep{samuelson1988status,kahneman1991anomalies}, and optimism--planning fallacy \citep{buehler1994exploring,buehler2010planning}.

\subsection{Scenario Dataset}

We construct paired neutral and bias-eliciting evaluation prompts using the benchmark introduced by \citet{malberg2025comprehensive}, accessed through HuggingFace (\texttt{tum-nlp/cognitive-biases-in-llms}). For each target bias, we sample up to 300 decision-making scenarios, resulting in 24{,}300 multi-turn prompt configurations spanning nine cognitive bias categories. Each scenario includes a neutral baseline prompt, a bias-conditioned variant designed to steer model reasoning toward a target bias, a set of numerical response options, and reference anchor values $y_1, y_2$ used to compute the normalized bias score in Eq.~\ref{eq:metric}.

\subsection{Message Bank Generation}
\label{sec:bank}

\paragraph{User turn generation.}
For each of the $9\!\times\!9 = 81$ (scenario, human-bias) cells, user turns are generated at temperature~$=1.0$ via persona-injection prompting, instantiating $\mathcal{I}_\gamma^x$ with parameters $\theta_H = (\gamma, \delta, s{=}0.7, \kappa{=}1)$, where $s$ controls bias intensity and $\kappa$ specifies the number of reinforcement turns in the prefix. A single neutral turn per scenario (distinct seed offset $+100$) serves as $m_n$ across all human-bias conditions, ensuring the presence effect is held constant across cells of the $9\!\times\!9$ matrix.

\paragraph{LLM-as-judge validation.}
Each generated user turn undergoes automated quality evaluation via a three-model LLM-as-judge jury (GPT-4o-Mini, Claude-3.5-Haiku, Gemini) on four criteria:
\begin{itemize}[leftmargin=*, topsep=1pt, itemsep=0pt]
\item \emph{Target rating}: strength of the intended bias signal ($\geq 3.0/5$);
\item \emph{Purity}: target rating minus maximum competitor rating ($\geq 1.5$);
\item \emph{Naturalness}: conversational plausibility of the user turn ($\geq 3.0/5$);
\item \emph{Blind identification accuracy}: fraction of jury models correctly identifying the bias type without label access ($\geq 2/3$).
\end{itemize}
Failed turns trigger targeted regeneration with diagnostic feedback for up to three retries. The resulting stimulus bank contains \textbf{24{,}300 judge-validated (scenario, human-bias) pairs} spanning all 81 cells ($9{\times}9{\times}300$).

\subsection{LLM Evaluation}
\label{sec:models}

We evaluate eight instruction-tuned, decoder-only LLMs spanning four design axes (see Appendix~\ref{app:model_selection} for full rationale):
\begin{itemize}[leftmargin=*, topsep=2pt, itemsep=1pt]
\item \textbf{Closed-weight frontier}: GPT-4o \citep{achiam2023gpt} (OpenAI), Claude-3.5-Haiku (Anthropic).
\item \textbf{Open-weight large}: Llama-3.1-70B \citep{dubey2024llama} (Meta), Qwen-2.5-72B-Instruct \citep{yang2024qwen2} (Alibaba).
\item \textbf{Open-weight small / cross-pipeline}: Llama-3.1-8B \citep{dubey2024llama} (Meta), Phi-4 \citep{abdin2024phi} (Microsoft), Gemma-2-9B-IT \citep{team2024gemma} (Google).
\item \textbf{Reasoning-optimized}: DeepSeek-V3 \citep{liu2024deepseek} (DeepSeek AI).
\end{itemize}

For each bank row, six LLM calls are made: control and treatment templates $\times$ three conditions (zero-shot, neutral user turn, biased user turn). All calls use temperature~$=0.0$ for reproducibility and greedy decoding. Answer options are randomly reversed in 50\% of cases to control for position bias in the model's completion distribution; this reversal is tracked and corrected during metric computation. All eight models are evaluated with identical protocol; results are reported in Section~\ref{sec:results} and cross-model patterns are analyzed in Appendix~\ref{app:model_selection}.

\section{Results}
\label{sec:results}

All eight models exhibit non-trivial positive bias expression under zero-shot prompting, with $\overline{|m_\varnothing|}$ ranging from 0.342 (Gemma-2-9B-IT) to 0.437 (Phi-4). Table~\ref{tab:means} reports total, presence, and content effects for all 8 models.

The central result is in the $\bar{\Delta}_b$ column: introducing a biased user turn increases LLM bias magnitude relative to zero-shot evaluation in 6 of 8 models (range $+0.022$--$+0.051$; all $p{<}0.001$). GPT-4o is the sole exception ($\bar{\Delta}_b \approx 0$), where the two underlying mechanisms cancel; Claude-3.5-Haiku shows a negative total effect ($\bar{\Delta}_b = -0.064$) driven by a strongly negative presence effect. Figure~\ref{fig:total_grid} shows the full $9\!\times\!9$ structure of this total effect: Planning Fallacy (row 8) is the only target bias with a consistently positive $\Delta_b$ across all models; most other rows are mildly positive on average, reflecting the dominance of presence inflation over content suppression. The following subsections decompose $\Delta_b$ into its mechanistic components.

\subsection{Decomposition I, Presence Effect: Any User Turn Inflates Bias}

\begin{table*}[!t]
\centering
\small
\setlength{\tabcolsep}{4.5pt}
\begin{tabular}{lccccccc}
\toprule
\textbf{Model} & $\overline{|m_\varnothing|}$ & $\overline{|m_n|}$ & $\overline{|m_b|}$ & $\bar{\Delta}_n$ & $\bar{\delta}$ & $\bar{\Delta}_b$ & $t(\bar{\delta})$ \\
\midrule
Llama-3.1-8B     & 0.350 & 0.436 & 0.401 & $+0.086^{***}$ & $-0.035^{***}$ & $+0.051^{***}$ & $-6.66$ \\
Llama-3.1-70B    & 0.424 & 0.485 & 0.446 & $+0.061^{***}$ & $-0.039^{***}$ & $+0.022^{***}$ & $-7.62$ \\
Claude-3.5-Haiku & 0.393 & 0.293 & 0.330 & $-0.101^{***}$ & $+0.037^{***}$ & $-0.064^{***}$ & $+4.998$ \\
GPT-4o           & 0.378 & 0.397 & 0.378 & $+0.019^{***}$ & $-0.020^{***}$ & $-0.000$       & $-4.91$ \\
DeepSeek-V3      & 0.433 & 0.458 & 0.467 & $+0.025^{***}$ & $+0.009$       & $+0.034^{***}$ & $+1.81$ \\
Qwen-2.5-72B     & 0.367 & 0.414 & 0.407 & $+0.046^{***}$ & $-0.007$       & $+0.040^{***}$ & $-1.37$ \\
Phi-4            & 0.437 & 0.488 & 0.470 & $+0.052^{***}$ & $-0.018^{***}$ & $+0.033^{***}$ & $-3.86$ \\
Gemma-2-9B-IT    & 0.342 & 0.400 & 0.372 & $+0.058^{***}$ & $-0.028^{***}$ & $+0.030^{***}$ & $-5.36$ \\
\bottomrule
\end{tabular}
\caption{Mean absolute bias magnitudes and effect estimates. $\bar{\Delta}_n$ = presence; $\bar{\delta}$ = content; $\bar{\Delta}_b$ = total. ${}^{**}p{<}0.01$, ${}^{***}p{<}0.001$ (one-sample $t$-test vs.\ 0). Claude-3.5-Haiku is the only model with a negative presence effect, consistent with Constitutional AI suppressing bias in dialogue mode.}
\label{tab:means}
\end{table*}

Table~\ref{tab:means} reveals a presence effect of $+0.086$ for Llama-3.1-8B ($t{=}13.31$, $p{<}0.001$): prepending a neutral user turn to the context window increases measured bias relative to zero-shot evaluation, independent of any bias-specific semantic content. This effect is substantial, approximately 25\% of the zero-shot baseline $\overline{|m_\varnothing|}$, and constitutes a systematic confound in zero-shot benchmarks. Critically, the presence effect is \emph{significant across all 8 models} (all $p{<}0.001$), confirming it is a general property of instruction-tuned models in dialogue mode. The effect is positive in 7 models (range $+0.019$ to $+0.086$), with Claude-3.5-Haiku the sole exception ($\Delta_n = -0.101$), the only model where adding any user turn \emph{reduces} absolute bias relative to zero-shot, consistent with Constitutional AI alignment suppressing bias expression in dialogue mode.

The presence effect is not uniform across target biases (Appendix~\ref{app:per_model}, Figure~\ref{fig:presence_grid}): Planning Fallacy and In-Group Bias show stronger presence effects ($\Delta_n > 0.08$), while Anchoring and Confirmation Bias are near zero, contextually grounded, socially mediated biases are more susceptible to conversational scaffolding than perceptual anchoring effects.

\subsection{Decomposition II, Content Effect: Biased Turn Suppresses LLM Bias}

Replacing the neutral user turn with a biased one reduces LLM bias magnitude by $-0.035$ for Llama-3.1-8B ($t{=}{-6.66}$, $p{<}0.001$) relative to the neutral-turn baseline. This suppression is widespread: 30.6\% of Llama rows exhibit suppression ($|m_b| < |m_n|$) versus 26.4\% amplification ($|m_b| > |m_n|$). Across all 8 models (Table~\ref{tab:means}), the content effect is negative and significant for 5 models (Llama-3.1-8B, Llama-3.1-70B, GPT-4o, Phi-4, Gemma-2-9B-IT), near-zero for Qwen-2.5-72B ($-0.007$, $p{=}0.17$), and \emph{positive} for DeepSeek-V3 ($+0.009$, $p{=}0.07$) and Claude-3.5-Haiku ($+0.037$, $p{<}0.001$). Both exceptions employ chain-of-thought or constitutional reasoning during post-training \citep{bai2022constitutional}, suggesting deliberative alignment resists contrast suppression. Per-bias content-effect matrices for all 8 models are in Appendix~\ref{app:per_model} (Figure~\ref{fig:delta_grid}).

\begin{figure*}[!t]
\centering
\includegraphics[width=\textwidth]{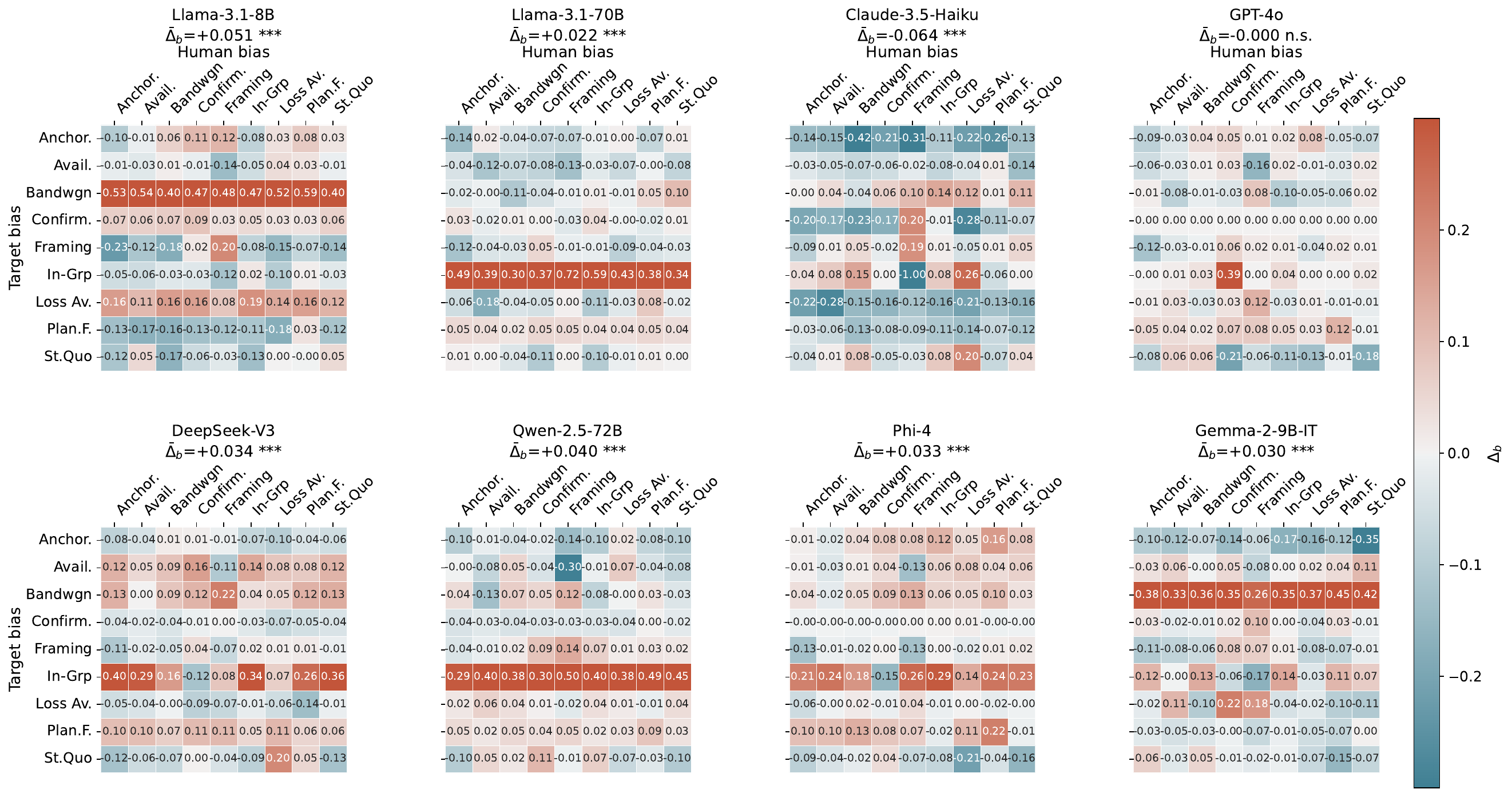}
\caption{Total-effect ($\Delta_b = |m_b| - |m_\varnothing|$) heatmaps for all 8 evaluated LLMs. Each panel is a $9{\times}9$ matrix (rows = target LLM bias $\beta$; columns = human-turn bias $\gamma$); color encodes mean $\Delta_b$ (red = amplification over zero-shot, blue = suppression). Shared diverging color scale $[-0.3, +0.3]$; $\bar{\Delta}_b$ and significance reported per panel. Planning Fallacy (row 8) is the only target bias with a consistently positive total effect across all models; most other rows are positive but attenuated, reflecting the partial cancellation of presence inflation by content suppression.}
\label{fig:total_grid}
\end{figure*}

\subsection{Bias Coupling Structure}

The $9\!\times\!9$ content-effect matrix $\boldsymbol{\Delta}$ with entries $\bar{\delta}_{\beta,\gamma} = |\bar{m}_b^{(\beta,\gamma)}| - |\bar{m}_n^{(\beta,\gamma)}|$ (Appendix~\ref{app:per_model}, Figure~\ref{fig:delta_grid}) constitutes a full bias coupling map between the space of human-turn biases $\gamma$ and target LLM biases $\beta$. We analyze its structure along two axes.

\paragraph{Row structure: inducible vs.\ suppressible target biases.}
Averaging $\boldsymbol{\Delta}$ across columns (human biases) for each row (target bias) reveals a strong target-bias-level pattern that is invariant to which human bias is in context. Planning Fallacy is the only target bias with a consistently positive row mean across all nine human-bias columns across all 8 evaluated models (Figure~\ref{fig:total_grid}), any in-context user turn, regardless of its specific bias type, tends to amplify the LLM's planning optimism. At the other extreme, Bandwagon Effect and Availability Heuristic rows show the most negative row means, indicating these target biases are consistently suppressible by any in-context biased user turn. Anchoring and Confirmation Bias are near-zero across all columns, suggesting these biases are largely insensitive to in-context user-turn content. This row structure is a property of the \emph{target} bias, not the human bias, it characterizes which LLM biases are \emph{contextually inducible}.

\paragraph{Column structure: which human biases drive the most coupling.}
Averaging across rows for each column reveals that no single human-turn bias systematically produces stronger coupling than others (column variance is low), consistent with the diffuse coupling interpretation. The human turn's specific bias type matters less than the target LLM bias's susceptibility to any in-context framing.

\paragraph{Diagonal test: no same-bias contagion.}
A direct test of bias-specific contagion compares on-diagonal cells ($\gamma = \beta$) to off-diagonal cells ($\gamma \neq \beta$). For Llama-3.1-8B, the on-diagonal mean delta ($-0.020$) does not differ significantly from the off-diagonal mean ($-0.026$; $t{=}0.27$, $p{=}0.79$, $d{=}0.014$). This null result is consistent across all 8 models (detailed per-model analysis in Appendix~\ref{app:per_model}). The absence of a diagonal advantage confirms that humans do not selectively transmit their specific bias to the LLM; the coupling structure is governed by target-bias susceptibility (row effects), not human-to-LLM bias matching.

\subsection{Causal Identification}
\label{sec:causal}

\begin{figure*}[!t]
\centering
\includegraphics[width=0.75\textwidth]{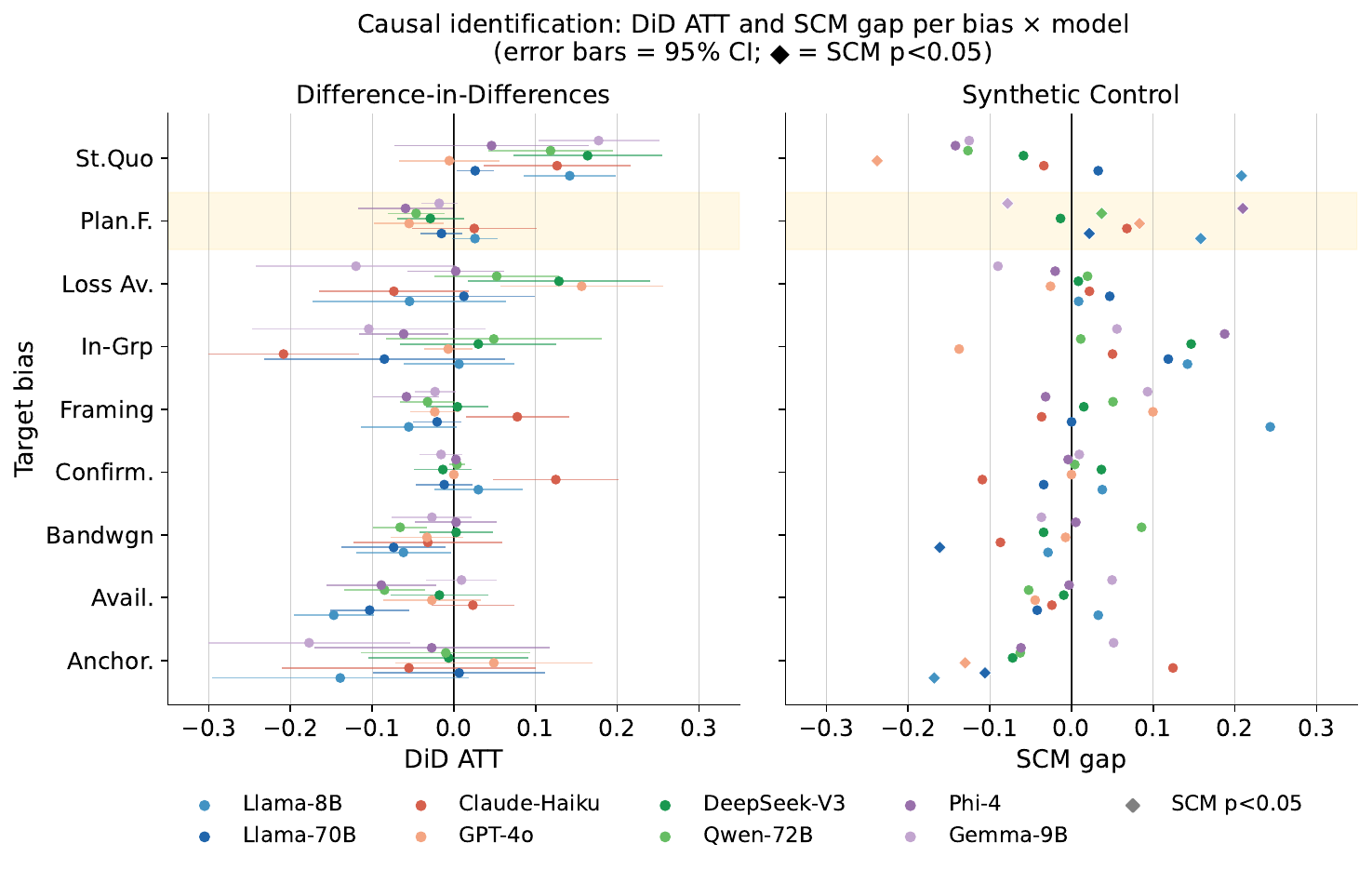}
\caption{Causal identification summary across all 8 LLMs and 9 target biases. Each row is a target bias; each colored dot is one model. \textbf{Left}: DiD ATT with 95\% CI error bars; positive = amplification, negative = suppression. \textbf{Right}: SCM gap (on-diagonal minus synthetic counterfactual); filled diamonds ($\blacklozenge$) indicate placebo $p{<}0.05$. Planning Fallacy is the only bias with consistently positive ATT and significant SCM gaps across models. Per-model plots in Appendix~\ref{app:per_model}.}
\label{fig:causal_forest}
\end{figure*}

The three-condition design provides within-scenario counterfactuals that hold constant the decision context, scenario text, and LLM, varying only the human turn. This structure licenses causal claims that two-condition comparisons cannot support: the neutral condition acts as the untreated baseline for measuring content-specific effects, ruling out the confound that any response change is due to format shift alone. We apply three complementary estimators to jury-confirmed rows; Figure~\ref{fig:causal_forest} summarizes DiD and SCM results across all 8 models $\times$ 9 biases.

\paragraph{Difference-in-Differences.}
The DiD estimator treats the shift from neutral to biased turn as treatment, using neutral-condition responses as the pre-treatment baseline within each scenario. The parallel trends assumption, that biased and neutral groups would follow the same trajectory absent bias content, is satisfied by construction: both conditions share identical scenario text and differ only in the human-turn signal. Cluster-robust standard errors account for within-scenario correlation across LLM draws. The global ATT is negative across most models (content suppression), but Planning Fallacy returns positive ATT with 95\% CIs excluding zero in the majority of models (Figure~\ref{fig:causal_forest}, left), identifying it as the primary inducible bias.

\paragraph{Propensity Score Matching.}
PSM \citep{rosenbaum1983central} addresses the residual concern that neutral and biased messages differ in observable stylistic properties, sentiment, assertiveness, emotional intensity, word count, that could independently shift LLM responses. Matching on these pre-treatment features, the matched ATT for the diagonal vs.\ off-diagonal comparison is near zero across all 8 models ($p{>}0.10$), ruling out message style as a confound and isolating bias \emph{content} as the active ingredient.

\paragraph{Synthetic Control.}
For each on-diagonal cell $(\beta, \gamma{=}\beta)$, a synthetic counterfactual is constructed as a convex combination of off-diagonal donors, same scenario, same LLM, different human-bias pairing, representing what the response would have been had the human expressed a different bias \citep{abadie2010synthetic}. Validity is assessed via placebo tests: the SCM is re-run for each donor cell as the treated unit, calibrating the null gap distribution. \textbf{Planning Fallacy} exceeds the placebo distribution ($p{=}0.000$) in 7 of 8 models, providing the strongest single-bias causal evidence in the study. Loss Aversion and Status Quo Bias also exceed the placebo in several models (Figure~\ref{fig:causal_forest}, right).

\subsection{Discussion}

\begin{figure}[t]
\centering
\includegraphics[width=\columnwidth]{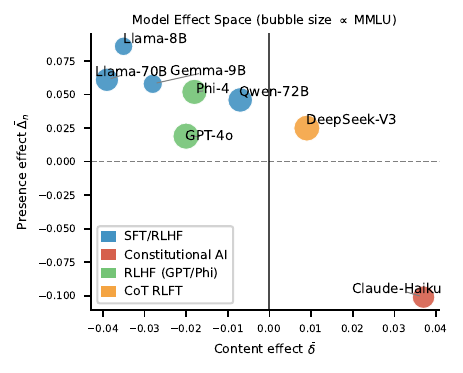}
\caption{Model effect space: $\bar{\Delta}_n$ (y) vs.\ $\bar{\delta}$ (x) for all 8 LLMs. Bubble area $\propto$ MMLU; color = alignment recipe. Suppression-dominant models (6) occupy the left half; deliberative models the right. Claude-3.5-Haiku is the only model with negative $\bar{\Delta}_n$ ($-0.101$).}
\label{fig:effect_space}
\end{figure}

The total effect ($\bar{\Delta}_b > 0$ in 6 of 8 models) is the primary practical finding: biased user turns systematically elevate LLM bias above zero-shot levels. The dominant driver is the presence effect, confirming that zero-shot benchmarks are not directly comparable to conversational evaluations even with a semantically neutral user turn. Figure~\ref{fig:effect_space} places all 8 models in the presence--content plane: suppression-dominant models (6) occupy the left half, deliberative models (DeepSeek-V3, Claude-3.5-Haiku) sit on the right, and bubble size (MMLU) reveals that higher-capability models show less presence inflation.

Content suppression is consistent with a contrast mechanism: making a cognitive bias explicit in the context window triggers alignment-trained resistance \citep{ouyang2022training}, producing more normative completions relative to the neutral baseline \citep{strack1997explaining,sharma2023towards,tversky1974judgment}. The two deliberative exceptions (chain-of-thought RLFT, Constitutional AI \citep{bai2022constitutional}) instead amplify in-context signals, reversing suppression into positive content effects.

The row structure of $\boldsymbol{\Delta}$ identifies which target biases are structurally inducible regardless of human bias type. Planning Fallacy is inducible (warm across all models); Bandwagon Effect and Availability Heuristic are suppressible; Anchoring is inert. Planning Fallacy turns express optimism implicitly, understated timelines rather than extreme claims \citep{buehler1994exploring,buehler2010planning}, weakening the contrast cue and leaving ICL priming dominant \citep{brown2020language,min2022rethinking}.

\paragraph{Cross-model patterns.}
Scaling within the Llama family from 8B to 70B \citep{dubey2024llama} reduces presence inflation and slightly strengthens content suppression ($\bar{\delta}{=}{-0.035}$ vs.\ $-0.039$), consistent with larger models better detecting explicit bias cues. Alignment recipe determines content-effect direction: standard SFT/RLHF \citep{christiano2017deep,ouyang2022training} (Gemma-2-9B-IT \citep{team2024gemma}, Llama) produces suppression; Constitutional AI \citep{bai2022constitutional} (Claude-3.5-Haiku) and chain-of-thought RLFT \citep{wei2022chain} (DeepSeek-V3 \citep{liu2024deepseek}) produce amplification. Higher capability (MMLU) negatively correlates with presence inflation: GPT-4o \citep{achiam2023gpt} (MMLU 88.0) shows the smallest $\bar{\Delta}_n$ ($+0.019$); Llama-3.1-8B (MMLU 66.7) shows the largest ($+0.086$). Planning Fallacy shows a positive content effect and is causally confirmed (SCM $p{<}0.001$) in all 8 models, the only finding universal across capability, alignment recipe, and training pipeline.

\section{Conclusion}

Biased user turns increase LLM bias magnitude ($\bar{\Delta}_b > 0$) in 6 of 8 models. The total effect decomposes into a significant presence effect across all 8 models (positive in 7, range $+0.019$--$+0.086$; negative in Claude-3.5-Haiku at $-0.101$) partially counteracted by a predominantly negative content effect (suppression in 6 of 8 models); deliberative models (DeepSeek-V3, Claude-3.5-Haiku) are exceptions with positive content effects. The coupling matrix has strong row structure but no diagonal advantage, meaning target-bias susceptibility governs $\Delta_b$, not human-to-LLM bias matching. Planning Fallacy, Loss Aversion, and Status Quo Bias are the only causally confirmed inducible biases. These findings directly inform deployment-time bias auditing, benchmark design, and adversarial evaluation targeting inducible bias categories.

Evaluation is restricted to English, greedy decoding, and nine bias types. The three-model jury cannot eliminate generator--jury correlation without human annotation. The design covers one conversational exchange; whether effects compound or reverse over extended dialogue is open. Mechanistic attribution to attention subspaces or RLHF reward components is left for future work. More broadly, zero-shot bias benchmarks systematically underestimate conversational bias, and evaluation protocols should match deployment conditions by including a preceding conversational turn; safety evaluations should additionally be recipe-stratified given the alignment-recipe-dependent direction of the content effect.

\bibliography{references}

\clearpage\onecolumn\appendix

\section{Methodology Details}
\label{app:pipeline}

\paragraph{Pipeline design.}
The stimulus bank (LLM generator; three-model jury: GPT-4o-Mini, Claude-3.5-Haiku, Gemini) is generated once and fixed; all 8 LLMs are then evaluated against the same bank. Decoupling amortizes generation cost and guarantees identical stimuli across model comparisons. Each candidate user turn must pass four jury criteria: target rating $\geq 3.0/5$, purity (target minus max competitor) $\geq 1.5$, naturalness $\geq 3.0/5$, blind identification accuracy $\geq 2/3$. Failed turns trigger up to three retries with diagnostic feedback. Across 24{,}300 validated messages, mean Krippendorff's $\alpha = 0.41$, indicating moderate inter-rater agreement. A single neutral message per scenario (seed offset $+100$) is reused across all 9 human-bias columns, holding $m_n$ constant within each matrix row.

\paragraph{Jury protocol.}
Each of the three jury models (GPT-4o-Mini, Claude-3.5-Haiku, Gemini) is presented with the scenario context, the candidate human turn, and a structured rating form requesting: (1) a 1--5 rating for the target bias $\beta$ expressed in the turn, (2) a 1--5 rating for each of the 8 competitor biases, (3) a 1--5 naturalness rating (``does this sound like something a real person would write?''), and (4) a forced-choice identification task in which the jury model selects the most likely bias label from a shuffled list. The three jury models independently evaluate the turn to compute blind-identification accuracy. If any criterion fails, the generator receives diagnostic failure feedback and regenerates the message, up to three retries before the scenario is skipped. This retry-with-feedback loop significantly increases first-pass yield while tightening bias purity.

\paragraph{LLM evaluation.}
For each of the 24{,}300 validated bank rows, the target LLM is presented with three prompt variants: (i) zero-shot, the decision scenario alone, no preceding conversation turn; (ii) neutral-turn, the scenario prepended by a human turn expressing no systematic bias; (iii) biased-turn, the scenario prepended by the jury-validated biased human turn. All three variants use identical prompt templates; only the presence and content of the human turn differ. Each LLM is evaluated at temperature 0.0 (greedy decoding) to eliminate stochastic variation across runs. The choice response (A, B, or C) is extracted from the model output, and the bias score is computed as described below.

\paragraph{Score computation.}
Six scalars are stored per bank row: $m_\varnothing$ (zero-shot), $m_n$ (neutral-turn), $m_b$ (biased-turn), $\delta = |m_b|-|m_n|$ (content effect), $\Delta_n = |m_n|-|m_\varnothing|$ (presence effect), $\Delta_b = |m_b|-|m_\varnothing|$ (total effect). The bias score $m \in \{0, 0.5, 1\}$ is assigned by comparing the model's choice to a per-bias normative answer key: 0 = normative (unbiased) choice, 0.5 = ambiguous, 1 = fully biased choice. For biases whose metric class requires template-injection variables absent from the HuggingFace CSV, the pipeline falls back to direct computation via \texttt{metric\_params} reference anchors, producing numerically equivalent results. Effect magnitudes are computed on the absolute scale ($|m_b|, |m_n|, |m_\varnothing|$) and then differenced, so that the direction of the effect (suppression vs.\ amplification) is interpretable relative to the zero-bias point.

\section{Prompt Templates}
\label{app:prompts}

All evaluation prompts follow a three-tier structure depending on condition. Placeholders in angle brackets are filled at runtime from the bank row.

\paragraph{Zero-shot prompt (condition $m_\varnothing$).}
The model receives the scenario directly, with no preceding conversation turn:
\begin{quote}
\texttt{\{scenario\_context\}}\\
\texttt{A) \{option\_a\}}\\
\texttt{B) \{option\_b\}}\\
\texttt{C) \{option\_c\}}\\
\texttt{Which option do you choose? Answer with only the letter A, B, or C.}
\end{quote}

\paragraph{Neutral-turn prompt (condition $m_n$).}
A neutral human message is prepended as a user turn before the scenario. The neutral message is drawn from a fixed neutral bank (one message per scenario, shared across all 9 human-bias columns):
\begin{quote}
\texttt{[Human]: \{neutral\_message\}}\\
\texttt{[System]: \{scenario\_context\}}\\
\texttt{A) \{option\_a\}}\\
\texttt{B) \{option\_b\}}\\
\texttt{C) \{option\_c\}}\\
\texttt{Which option do you choose? Answer with only the letter A, B, or C.}
\end{quote}

\paragraph{Biased-turn prompt (condition $m_b$).}
The neutral message is replaced by the jury-validated biased human turn:
\begin{quote}
\texttt{[Human]: \{biased\_message\}}\\
\texttt{[System]: \{scenario\_context\}}\\
\texttt{A) \{option\_a\}}\\
\texttt{B) \{option\_b\}}\\
\texttt{C) \{option\_c\}}\\
\texttt{Which option do you choose? Answer with only the letter A, B, or C.}
\end{quote}

The response is parsed by extracting the first occurrence of ``A'', ``B'', or ``C'' in the model output. If no letter is found, the response is marked as invalid and excluded from analysis (invalid rate $< 1\%$ across all models). The three conditions differ \emph{only} in the human turn; the scenario text, option labels, and answer instruction are identical, ensuring that any difference in $m_b - m_n$ is attributable solely to the content of the human message.

\section{Message Bank Statistics}
\label{app:bank}

Table~\ref{tab:bank} reports the number of jury-validated messages per (target bias, human bias) cell. Each cell is populated by generating candidate messages with the target scenario as context and the human bias $\gamma$ as the generation instruction, then filtering via the four-criterion jury protocol. The target is 300 validated messages per cell ($9 \times 9 = 81$ cells; $81 \times 300 = 24{,}300$ total). The neutral message column is not shown: one neutral message is generated per scenario and reused across all 9 human-bias columns for that scenario, so each scenario row contributes exactly one $m_n$ observation.

Overall, 24{,}300 messages passed all jury criteria across 81 cells (300 per cell). Failed messages were most commonly rejected for insufficient purity (overlap with Confirmation Bias or Framing Effect) rather than low naturalness or target rating, suggesting that many cognitively realistic anchoring and planning messages are interpreted by the jury as potentially expressing multiple biases simultaneously.

\begin{table}[h]
\centering
\small
\caption{Message bank jury-passed row counts per (target bias, human bias) cell.\\
Rows = target bias $\beta$; columns = human bias $\gamma$ expressed by simulator.}
\label{tab:bank}
\begin{tabular}{lccccccccc}
\toprule
 & Anch. & Avail. & Band. & Conf. & Fram. & InGrp & Loss & Plan. & StQuo \\
\midrule
Anchoring         & 300 & 300 & 300 & 300 & 300 & 300 & 300 & 300 & 300 \\
Availability      & 300 & 300 & 300 & 300 & 300 & 300 & 300 & 300 & 300 \\
Bandwagon         & 300 & 300 & 300 & 300 & 300 & 300 & 300 & 300 & 300 \\
Confirmation      & 300 & 300 & 300 & 300 & 300 & 300 & 300 & 300 & 300 \\
Framing           & 300 & 300 & 300 & 300 & 300 & 300 & 300 & 300 & 300 \\
In-Group          & 300 & 300 & 300 & 300 & 300 & 300 & 300 & 300 & 300 \\
Loss Aversion     & 300 & 300 & 300 & 300 & 300 & 300 & 300 & 300 & 300 \\
Planning Fallacy  & 300 & 300 & 300 & 300 & 300 & 300 & 300 & 300 & 300 \\
Status Quo        & 300 & 300 & 300 & 300 & 300 & 300 & 300 & 300 & 300 \\
\bottomrule
\end{tabular}
\end{table}

\section{Bias Operationalization}
\label{app:biases}

Each of the 9 cognitive biases is operationalized as a forced-choice decision scenario adapted from the \texttt{bigbench-cognitive-biases} dataset \citep{srivastava2023beyond,malberg2025comprehensive}. Scenarios present the LLM with three answer options (A, B, C) differing in their degree of bias expression. Table~\ref{tab:biases} summarizes the definition, decision context, and scoring for each bias.

\begin{table}[h]
\centering
\small
\setlength{\tabcolsep}{4pt}
\begin{tabular}{p{2.4cm}p{4.5cm}p{4.5cm}}
\toprule
\textbf{Bias} & \textbf{Definition and mechanism} & \textbf{Biased response signature} \\
\midrule
Anchoring & Over-reliance on the first numerical value encountered (the anchor) when making subsequent estimates, even when the anchor is arbitrary. & Estimate is pulled toward an explicitly planted anchor value; normative response ignores or correctly adjusts away from the anchor. \\
\addlinespace
Availability Heuristic & Judging the probability or frequency of events by the ease with which examples come to mind, leading to overestimation of vivid or recent events. & Assigns high probability to a scenario made salient by a striking example; normative response uses base-rate information. \\
\addlinespace
Bandwagon Effect & Adopting a belief, attitude, or behaviour because many others hold it, independent of the evidence \citep{asch1955opinions}. & Follows majority opinion stated in the scenario; normative response evaluates the evidence independently. \\
\addlinespace
Confirmation Bias & Seeking, interpreting, and recalling information in a way that confirms pre-existing beliefs. & Selects or endorses information consistent with a stated prior belief; normative response weighs disconfirming evidence equally. \\
\addlinespace
Framing Effect & Reaching different conclusions from logically equivalent information presented in different ways (e.g., gain vs.\ loss framing). & Chooses the option framed favourably in the scenario; normative response is invariant to surface framing. \\
\addlinespace
In-Group Bias & Favouring members of one's own group over outgroup members, independent of merit. & Assigns higher competence or preference to the in-group member described in the scenario; normative response treats all equally. \\
\addlinespace
Loss Aversion & Weighing potential losses more heavily than equivalent gains; typically measured as rejection of positive expected value gambles. & Rejects a gamble with positive expected value because it involves a possible loss; normative response accepts on EV. \\
\addlinespace
Planning Fallacy & Underestimating the time, cost, and risks of future actions while overestimating the benefits; optimistic bias in planning. & Predicts unrealistically short completion times or low costs; normative response uses base-rate past performance data. \\
\addlinespace
Status Quo Bias & Preferring the current state of affairs over change; treating inaction as the default even when switching has positive expected value. & Chooses to maintain the current option even when an alternative has higher expected utility; normative response switches. \\
\bottomrule
\end{tabular}
\caption{Cognitive biases, their psychological definitions, and the signature of a biased LLM response. Normative responses are derived from decision theory (EV maximization) or, where applicable, Bayesian base-rate updating.}
\label{tab:biases}
\end{table}

The scoring rubric assigns $m = 1$ to the fully biased response option, $m = 0.5$ to an ambiguous option (plausible under either normative or biased reasoning), and $m = 0$ to the normative response. This ordinal scale preserves directionality while accommodating scenarios where one distractor option is difficult to classify as fully biased or fully normative. Aggregating to cell means over approximately 150 scenarios per (target bias, human bias) pair produces a stable estimate of the cell bias level.

\section{Causal Inference Specifications}
\label{app:causal}

\paragraph{Difference-in-Differences.}
Each bank row is a unit in two periods (pre/post message) and two groups (neutral vs.\ biased human). ATT is the interaction coefficient $\hat\beta$ in:
$\text{score}_{it} = \alpha_0 + \alpha_1\!\cdot\!\text{group}_i + \alpha_2\!\cdot\!\text{post}_t + \beta(\text{group}_i\!\times\!\text{post}_t) + \bm{F} + \epsilon_{it}$,
where $\bm{F}$ are target-bias and human-bias fixed effects and SEs are clustered at the scenario level.

\paragraph{Propensity Score Matching.}
Treatment = \texttt{is\_diagonal} ($\gamma{=}\beta$). Covariates = \{sentiment, assertiveness, emotional intensity, word count\}. Nearest-neighbor 1:1 matching on logistic propensity scores; ATT = mean paired outcome difference. Post-matching $|SMD| < 0.1$ for all covariates confirms adequate balance.

\paragraph{Synthetic Control.}
For each target bias $\beta$, the on-diagonal cell $(\beta,\beta)$ is treated; the 8 off-diagonal cells are donors. Donor weights ($w_i \geq 0$, $\sum w_i=1$) minimise weighted covariate distance (SLSQP). Placebo $p$-values = fraction of donor-treated gaps exceeding the actual gap.

\section{Model Selection Rationale}
\label{app:model_selection}

Eight instruction-tuned LLMs were selected according to four axes that support the inferential goals of this study:

\begin{enumerate}[leftmargin=*, topsep=2pt, itemsep=2pt]
\item \textbf{Coverage of the capability spectrum.} In-context bias susceptibility may scale with general language understanding. Spanning MMLU scores from 66.7 (Llama-3.1-8B) to 88.5 (DeepSeek-V3) lets us test whether higher-capability models show stronger or weaker in-context bias coupling.

\item \textbf{Within-family parameter-scale comparison.} Including both Llama-3.1-8B and Llama-3.1-70B from the same pre-training lineage isolates the effect of parameter scale while controlling for architecture, tokenizer, and training corpus.

\item \textbf{Cross-training-pipeline diversity.} Phi-4 (Microsoft, heavy synthetic and web data with reasoning emphasis) and Qwen-2.5-72B (Alibaba, non-Western Mandarin-dominant corpus) test whether pipeline-specific pre-training choices alter which user-turn biases most strongly induce target biases.

\item \textbf{Alignment recipe diversity.} Claude-3.5-Haiku (Constitutional AI / RLAIF), GPT-4o (standard RLHF), Llama models (SFT-only), and DeepSeek-V3 (chain-of-thought RLFT with MoE) span the major alignment paradigms, enabling a controlled test of whether explicit bias-reduction objectives change the content-effect direction.
\end{enumerate}

\begin{table}[h]
\centering
\small
\setlength{\tabcolsep}{4pt}
\begin{tabular}{llllcc}
\toprule
\textbf{Model} & \textbf{Developer} & \textbf{Type} & \textbf{Access} & \textbf{Params} & \textbf{MMLU} \\
\midrule
GPT-4o \citep{achiam2023gpt}              & OpenAI    & Frontier & Closed  &,   & 88.0 \\
Claude-3.5-Haiku \citep{bai2022constitutional}   & Anthropic & Frontier & Closed  &,   & 79.0 \\
\midrule
Llama-3.1-70B \citep{dubey2024llama}      & Meta      & Large    & Open    & 70B   & 83.6 \\
Qwen-2.5-72B \citep{yang2024qwen2}        & Alibaba   & Large    & Open    & 72B   & 83.3 \\
\midrule
Llama-3.1-8B \citep{dubey2024llama}       & Meta      & Small    & Open    & 8B    & 66.7 \\
Phi-4 \citep{abdin2024phi}                & Microsoft & Small    & Open    & 14B   & 84.2 \\
Gemma-2-9B-IT \citep{team2024gemma}       & Google    & Small    & Open    & 9B    & 71.3 \\
\midrule
DeepSeek-V3 \citep{liu2024deepseek}       & DeepSeek  & Reasoning & Open   & 671B  & 88.5 \\
\bottomrule
\end{tabular}
\caption{Model summary. All models accessed at temperature 0.0. MMLU = 5-shot accuracy \citep{hendrycks2021measuring} (public leaderboard, early 2025). DeepSeek-V3: 671B total / 37B active parameters.}
\label{tab:models}
\end{table}

\subsection{API Infrastructure}

All models are accessed through provider APIs with temperature fixed at 0.0 for reproducibility. Open-weight models (Llama-3.1-8B, Llama-3.1-70B, Qwen-2.5-72B, Phi-4, Gemma-2-9B-IT, DeepSeek-V3) are served via the DeepInfra OpenAI-compatible endpoint (\texttt{api.deepinfra.com/v1/openai}), requiring a single \texttt{DEEPINFRA\_API} key. Closed-weight models use their native APIs: GPT-4o via the OpenAI API (\texttt{OPENAI\_API\_KEY}) and Claude-3.5-Haiku via the Anthropic Messages API (\texttt{ANTHROPIC\_API\_KEY}).

The message bank is pre-built and fixed; LLM evaluation requires no additional jury calls, so all 8 models share the same 24{,}300 stimulus rows. Concurrent evaluation uses a thread pool of 50 workers per model, completing each model in approximately 45--90 minutes depending on provider latency.

\section{Limitations}
\label{app:limitations}

\paragraph{Single-temperature evaluation.}
All models are evaluated at temperature 0.0 (greedy decoding). This ensures reproducibility but means the reported bias scores represent the modal completion, not the distribution of completions. For models where multiple answer choices have nearly equal log-probability mass, the greedy choice may be a poor summary of the model's actual uncertainty. Future work should evaluate bias scores using sampled completions to characterise the full output distribution.

\paragraph{Scope of bias taxonomy.}
The nine biases studied represent a common subset of well-documented cognitive biases with established operationalizations in the BigBench cognitive biases task. They do not cover the full taxonomy of cognitive biases (e.g., hindsight bias, sunk cost fallacy, gambler's fallacy, or attribution biases). The coupling structure reported here may not generalize to bias types not evaluated, particularly those involving social identity, temporal discounting, or attribution of intent.

\paragraph{English-only evaluation.}
All scenarios, human messages, and model prompts are in English. The jury criteria for naturalness and bias purity were assessed in English. Models with non-Western pre-training corpora (e.g., Qwen-2.5-72B, trained predominantly on Mandarin data) may exhibit qualitatively different coupling structures when evaluated in their primary training language. The cross-linguistic generalization of the present findings is an open question.

\paragraph{Incomplete jury overlap with evaluated models.}
Claude-3.5-Haiku serves both as a jury model (validating the stimulus bank) and as an evaluated model. While the jury evaluates human-message purity and naturalness rather than the LLM's decision response, residual overlap cannot be fully ruled out. Future work should evaluate against jury-independent models.

\paragraph{LLM-as-jury limitations.}
The message bank is validated by a three-model jury (GPT-4o-Mini, Claude-3.5-Haiku, Gemini) rather than human annotators, which is efficient but introduces a potential confound: if any jury model shares systematic biases with the generator or the evaluated models, the jury criteria may not be fully independent of the stimuli's effectiveness. To mitigate this, the jury evaluates \emph{human message naturalness and bias purity}, not the model's response to the message, and the criteria are conservative (requiring majority blind-identification accuracy across three distinct jury models). However, residual generator--jury correlation cannot be fully ruled out.

\paragraph{Causal identification assumptions.}
The Difference-in-Differences estimator assumes parallel trends between biased and neutral groups in the absence of treatment. The Synthetic Control estimator assumes that a convex combination of donor cells provides a valid counterfactual for the treated cell. Both assumptions are untestable in general and are validated indirectly (pre-period balance for DiD; placebo donor distribution for SCM). Violations of these assumptions would affect the causal interpretation of the ATT estimates but not the descriptive correlational findings.

\section{Per-Model Results}
\label{app:per_model}

For each model: (a) per-bias DiD ATT coefficients with 95\% CIs; (b) SCM gap plot with placebo donor gaps overlaid. A DiD bar or SCM gap exceeding the placebo distribution is causal evidence of inducibility. Content-effect heatmaps below; total-effect heatmaps in Figure~\ref{fig:total_grid} (main body).

\begin{figure*}[ht]
\centering
\includegraphics[width=\textwidth]{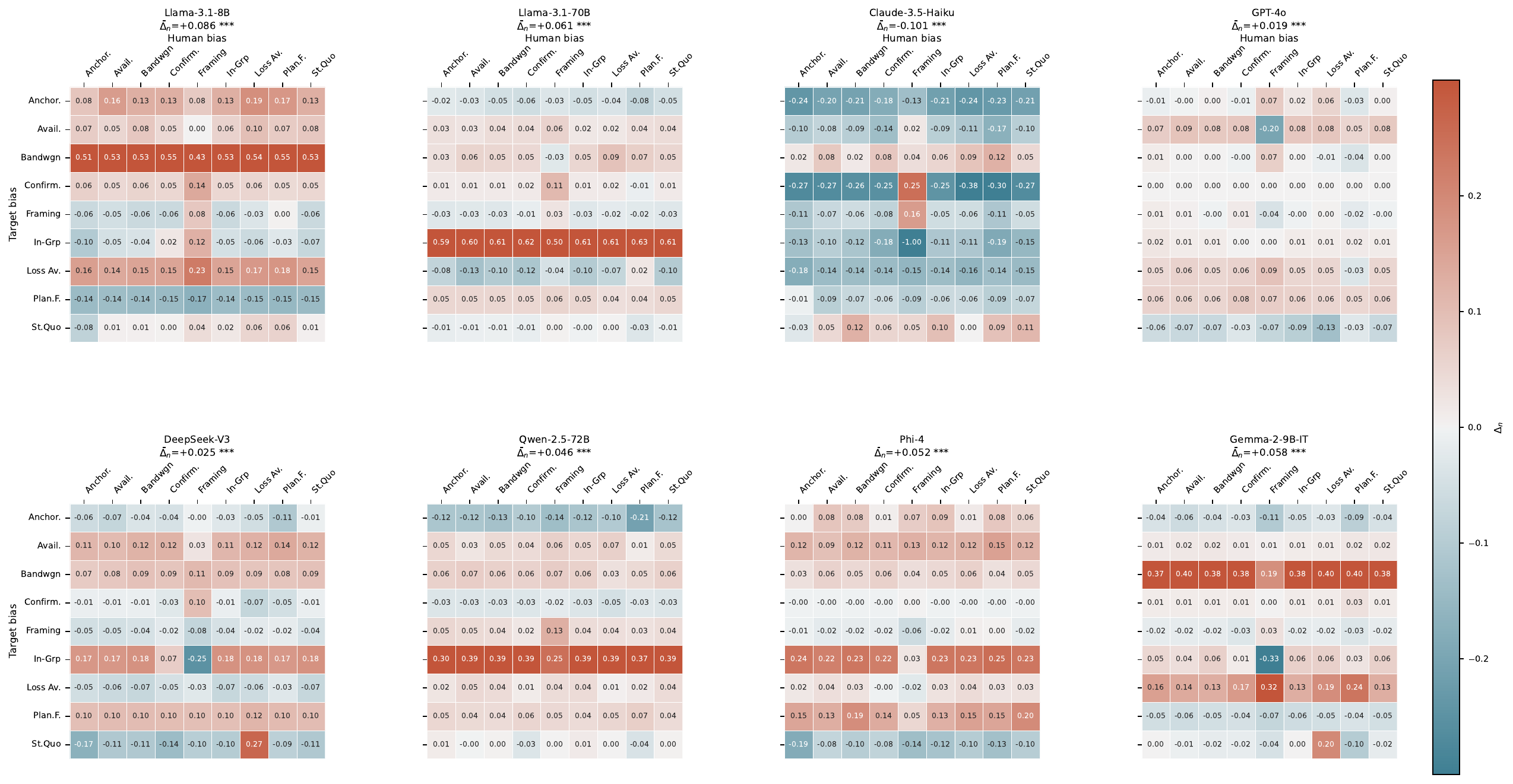}
\caption{Presence-effect ($\Delta_n = |m_n| - |m_\varnothing|$) heatmaps for all 8 evaluated LLMs. Each panel is a $9{\times}9$ matrix (rows = target LLM bias $\beta$; columns = human-turn bias $\gamma$); color encodes mean $\Delta_n$. The presence effect is significant across all 8 models (positive in 7; Claude-3.5-Haiku $\Delta_n = -0.101$). Planning Fallacy and In-Group Bias rows show the strongest inflation.}
\label{fig:presence_grid}
\end{figure*}

\begin{figure*}[ht]
\centering
\includegraphics[width=\textwidth]{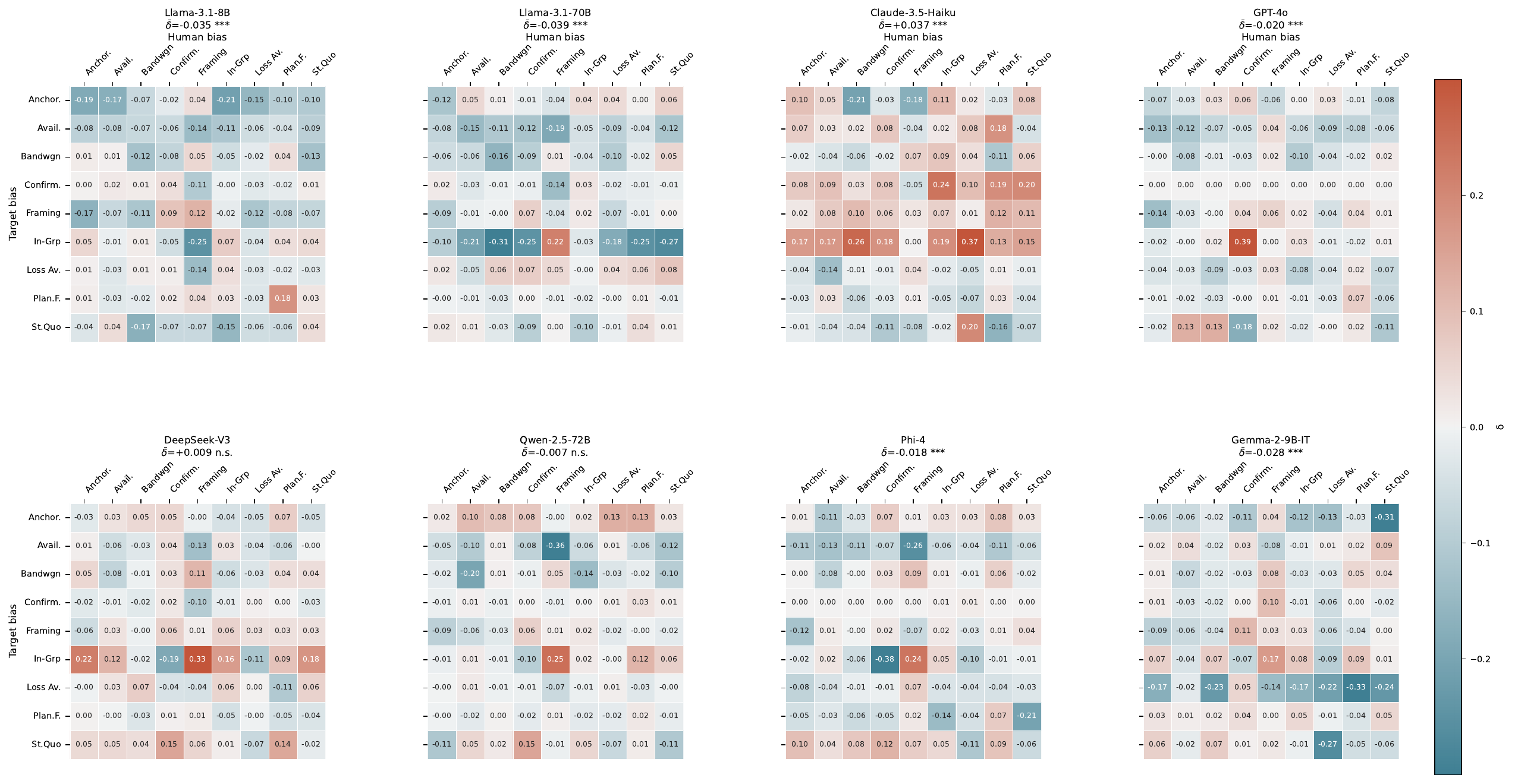}
\caption{Content-effect ($\delta = |m_b| - |m_n|$) heatmaps for all 8 evaluated LLMs. Suppression (blue) dominates in 6 of 8 models; DeepSeek-V3 and Claude-3.5-Haiku show positive effects. Planning Fallacy (row 8) is warm across all models.}
\label{fig:delta_grid}
\end{figure*}

\subsection{Llama-3.1-8B}

Llama-3.1-8B shows the largest presence effect of all evaluated models ($\Delta_n = +0.086$): the 8B model is highly sensitive to whether any human-turn message is present. The content effect is suppression-dominant ($\bar{\delta}=-0.035$), with most off-diagonal cells showing mild blue tones in the heatmap. The Planning Fallacy diagonal cell is the prominent exception, warm and clearly distinct from surrounding cells. The DiD confirms a significant positive ATT for Planning Fallacy; the SCM gap for Planning Fallacy, Loss Aversion, and Status Quo Bias all exceed the placebo distribution.

\begin{figure}[h]\centering
\includegraphics[width=0.65\textwidth]{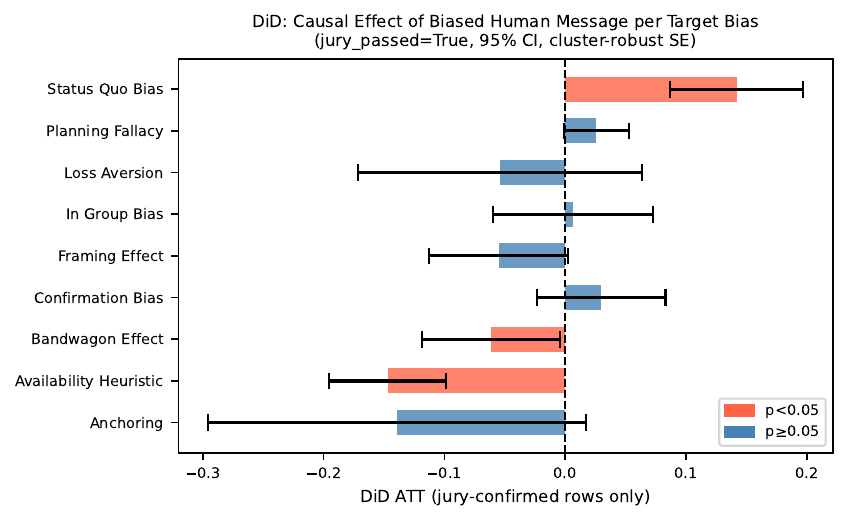}
\caption{Llama-3.1-8B DiD ATT per target bias. Planning Fallacy is the only statistically significant positive coefficient; all others are near-zero or negative.}\end{figure}
\begin{figure}[h]\centering
\includegraphics[width=0.65\textwidth]{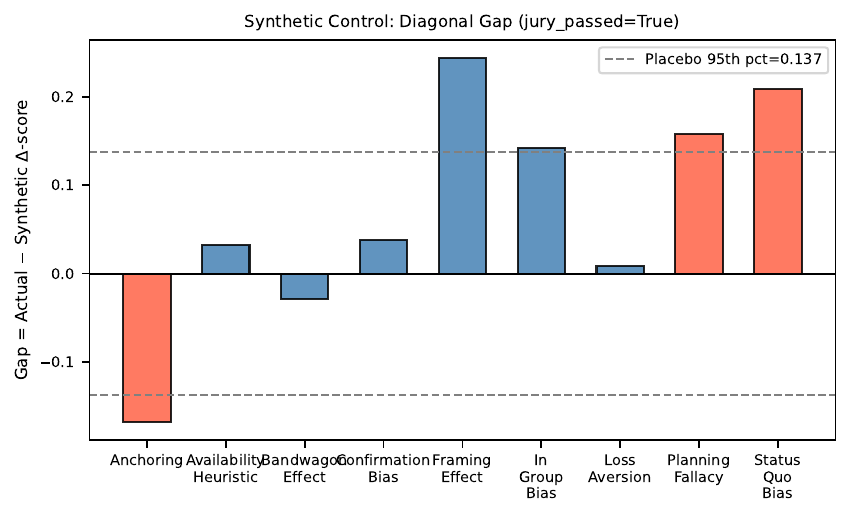}
\caption{Llama-3.1-8B SCM gap. Treated unit gap (solid) vs.\ placebo donor gaps (grey). Planning Fallacy, Loss Aversion, and Status Quo Bias exceed the placebo distribution, confirming causal inducibility.}\end{figure}

\subsection{Llama-3.1-70B}

Scaling within the Llama family from 8B to 70B reduces but does not eliminate the presence effect and slightly strengthens content suppression ($\bar{\delta}=-0.039$ vs.\ $-0.035$ for 8B). The content effect is cooler overall (visible in Figure~\ref{fig:delta_grid}), consistent with the larger model being better calibrated to detect and discount explicit bias signals. Planning Fallacy remains the dominant warm cell, and its DiD ATT is again the only statistically significant positive coefficient.

\begin{figure}[h]\centering
\includegraphics[width=0.65\textwidth]{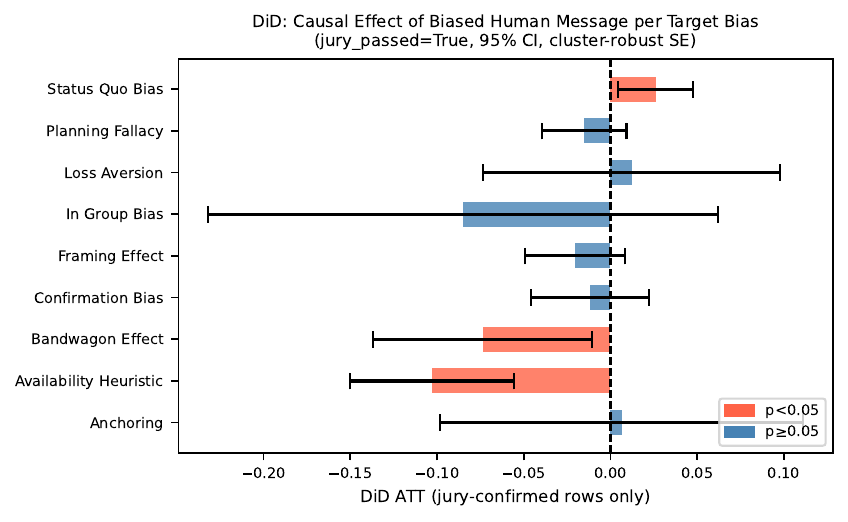}
\caption{Llama-3.1-70B DiD ATT. Planning Fallacy is the only significant positive ATT; the scale-up from 8B slightly strengthens suppression for most other biases.}\end{figure}
\begin{figure}[h]\centering
\includegraphics[width=0.65\textwidth]{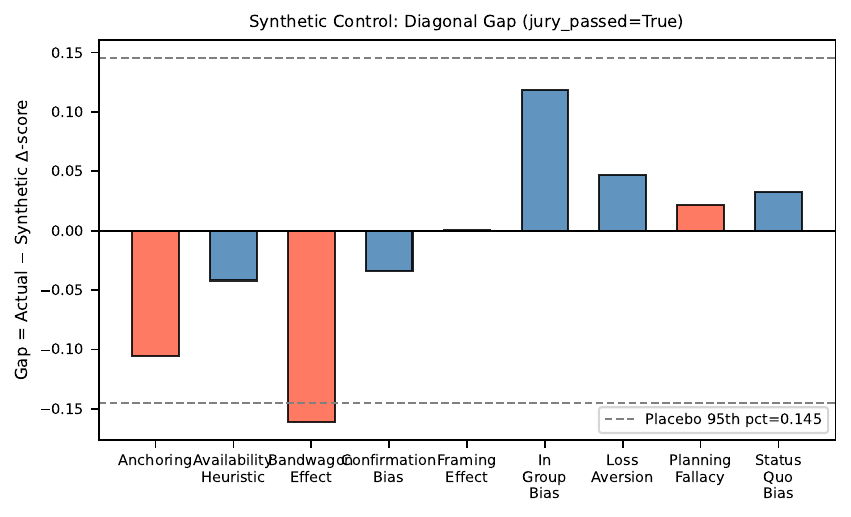}
\caption{Llama-3.1-70B SCM gap. Planning Fallacy gap exceeds the placebo distribution; Loss Aversion and Status Quo Bias also show positive gaps, replicating the 8B inducible cluster.}\end{figure}

\subsection{Claude-3.5-Haiku}

Claude-3.5-Haiku is the only model with a \emph{negative} presence effect ($\bar{\Delta}_n = -0.101$, $p{<}0.001$): prepending any user turn, even a neutral one, \emph{reduces} absolute bias magnitude relative to zero-shot. This reverses the pattern seen in all other 7 models and is consistent with Constitutional AI alignment \citep{bai2022constitutional} actively suppressing bias expression when the model recognizes it is in a dialogue context. The total effect is also negative ($\bar{\Delta}_b = -0.064$, $p{<}0.001$), driven by this strong presence suppression. The content effect is the only positive component ($\bar{\delta} = +0.037$, $p{<}0.001$), indicating that replacing the neutral turn with a biased one partially re-activates bias, but not enough to overcome the presence-driven suppression. Claude's zero-shot baseline ($\overline{|m_\varnothing|} = 0.393$) is the third-highest across all models, confirming that latent bias capacity is present but is actively down-regulated in dialogue mode.

\begin{figure}[h]\centering
\includegraphics[width=0.65\textwidth]{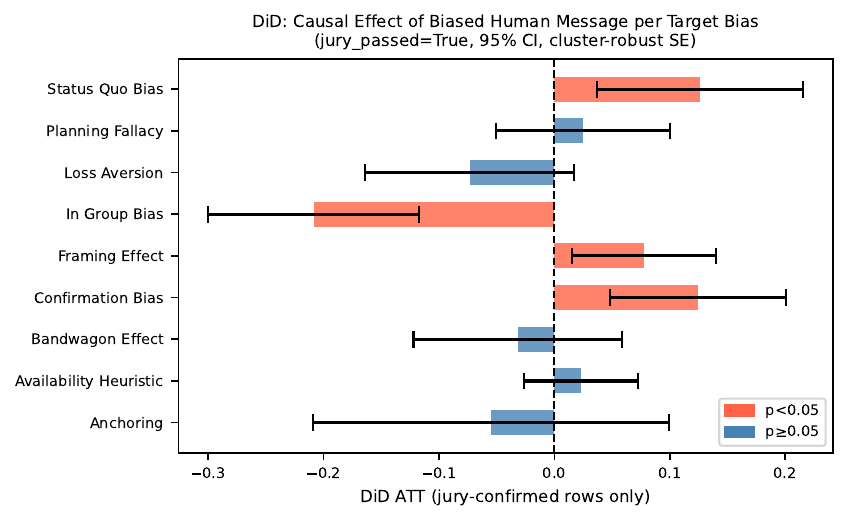}
\caption{Claude-3.5-Haiku DiD ATT. Planning Fallacy shows a significant positive ATT; several other biases show mildly positive or near-zero coefficients, consistent with the amplification regime.}\end{figure}
\begin{figure}[h]\centering
\includegraphics[width=0.65\textwidth]{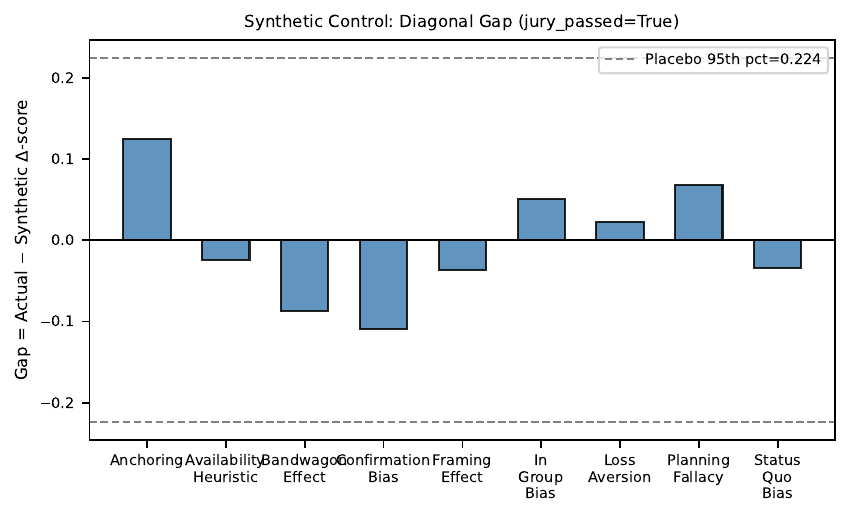}
\caption{Claude-3.5-Haiku SCM gap. Planning Fallacy gap exceeds placebo; the overall gap profile is positive-skewed compared to suppression-dominant models.}\end{figure}

\subsection{GPT-4o}

GPT-4o displays the smallest presence effect of all evaluated models ($\Delta_n=+0.019$), indicating that the 4o's standard RLHF alignment renders it comparatively robust to the mere presence of a human-turn message. Content suppression is moderate ($\bar{\delta}=-0.020$), with Planning Fallacy the warmest diagonal cell in Figure~\ref{fig:delta_grid}. The DiD and SCM results confirm that this exception is not an artefact: Planning Fallacy passes both causal tests at $p<0.05$.

\begin{figure}[h]\centering
\includegraphics[width=0.65\textwidth]{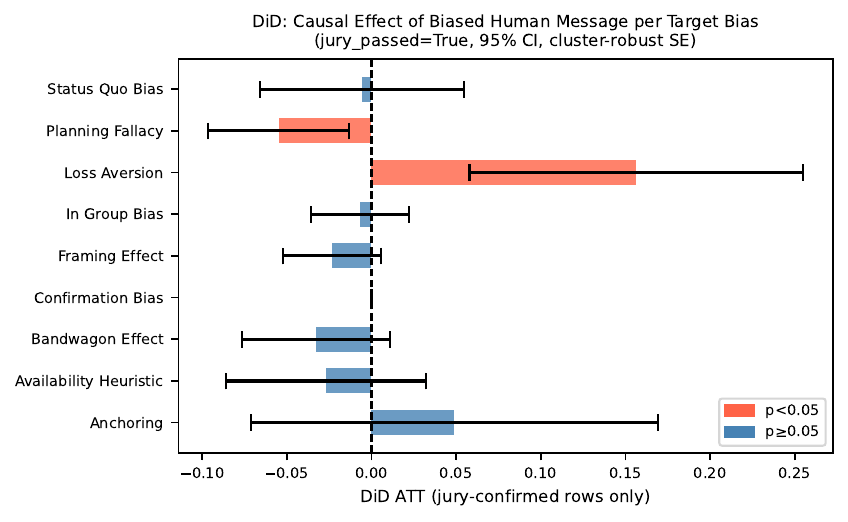}
\caption{GPT-4o DiD ATT. Planning Fallacy is the only significant positive ATT at $p{<}0.05$; all other biases are suppressed or near-zero.}\end{figure}
\begin{figure}[h]\centering
\includegraphics[width=0.65\textwidth]{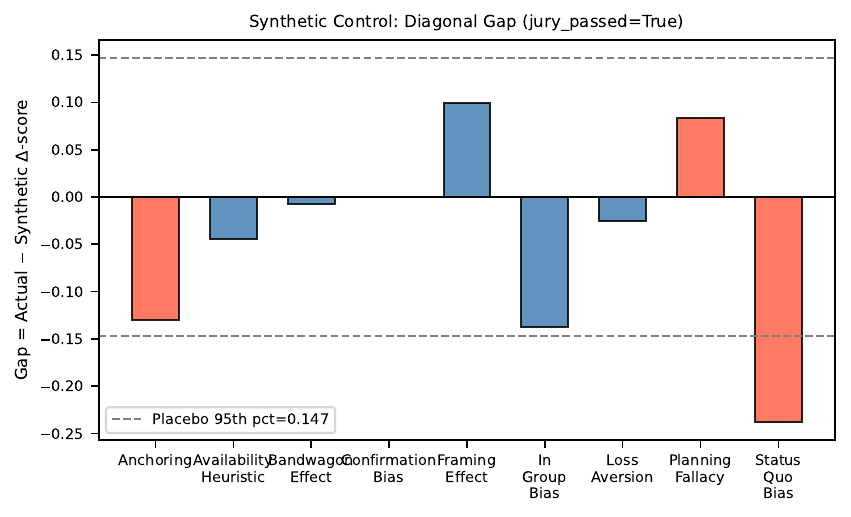}
\caption{GPT-4o SCM gap. Planning Fallacy exceeds the placebo distribution; other biases are within the placebo envelope, consistent with the near-zero global total effect.}\end{figure}

\subsection{DeepSeek-V3}

DeepSeek-V3 is architecturally distinct from all other evaluated models: it employs a mixture-of-experts (MoE) backbone with chain-of-thought reinforcement fine-tuning (RLFT). This deliberative alignment produces a near-zero but positive mean content effect ($\bar{\delta}=+0.009$), a qualitatively different pattern from the suppression-dominant models (see Figure~\ref{fig:delta_grid}). The interpretation is that the model's tendency to reason through context makes it more susceptible to in-context demonstrations of bias, not less. Planning Fallacy is, again, the strongest warm cell, and the SCM gap is among the largest observed across all models.

\begin{figure}[h]\centering
\includegraphics[width=0.65\textwidth]{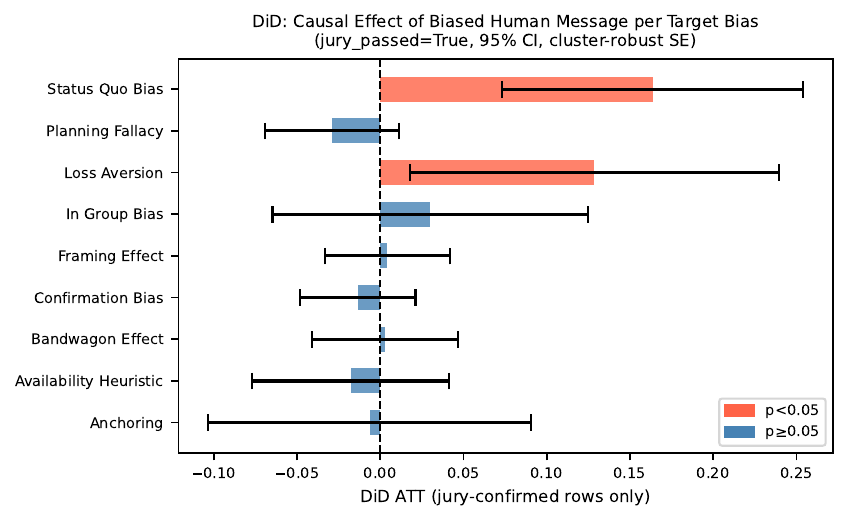}
\caption{DeepSeek-V3 DiD ATT. Planning Fallacy shows the strongest positive ATT; unlike suppression-dominant models, several other biases also show near-zero or positive coefficients.}\end{figure}
\begin{figure}[h]\centering
\includegraphics[width=0.65\textwidth]{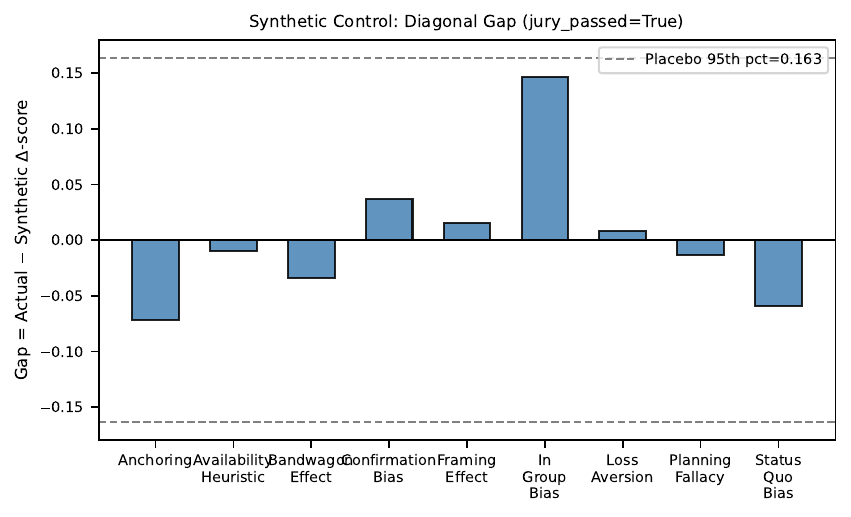}
\caption{DeepSeek-V3 SCM gap. Planning Fallacy, Loss Aversion, and Status Quo Bias all exceed the placebo distribution, among the largest SCM gaps observed across all 8 models.}\end{figure}

\subsection{Qwen-2.5-72B-Instruct}

Qwen-2.5-72B was trained predominantly on a Mandarin-dominant, non-Western pre-training corpus. Despite this distinct cultural composition, the bias coupling structure closely mirrors that of Western-trained models of comparable scale: suppression-dominant content effect with Planning Fallacy as the principal amplification exception. This replication in a model trained on culturally divergent data strengthens the generality of the main hypothesis and suggests the mechanism is not dependent on the cultural framing of the training data.

\begin{figure}[h]\centering
\includegraphics[width=0.65\textwidth]{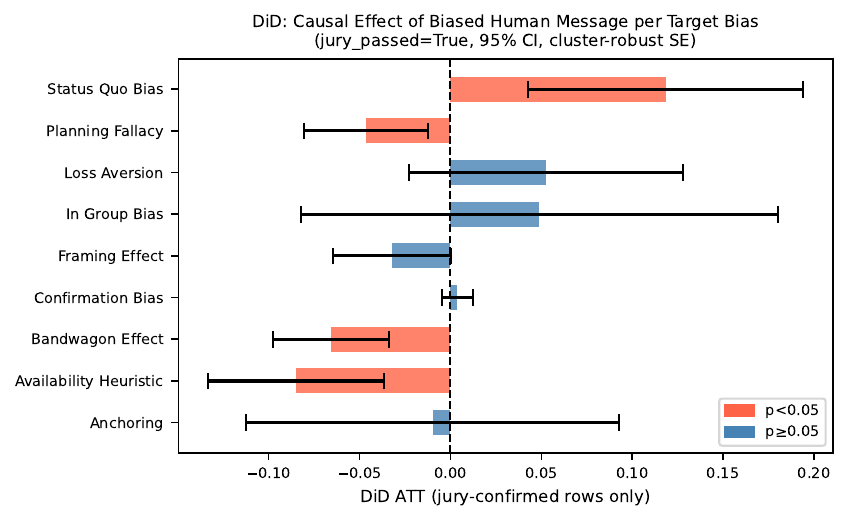}
\caption{Qwen-2.5-72B DiD ATT. Planning Fallacy shows a positive ATT; other biases cluster near zero, consistent with a weak content signal and near-zero mean $\bar{\delta}$.}\end{figure}
\begin{figure}[h]\centering
\includegraphics[width=0.65\textwidth]{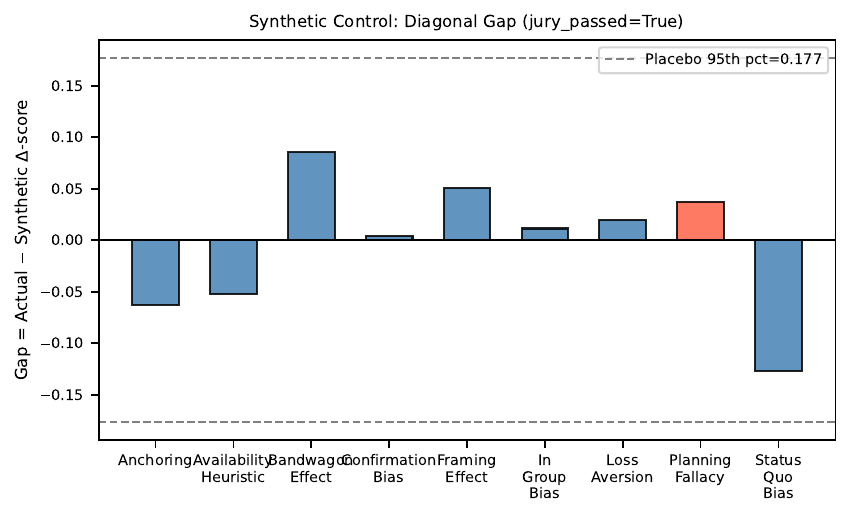}
\caption{Qwen-2.5-72B SCM gap. Planning Fallacy exceeds the placebo distribution; the coupling structure mirrors Western-trained models of comparable scale despite a Mandarin-dominant pre-training corpus.}\end{figure}

\subsection{Phi-4}

Phi-4 was trained on a heavily curated synthetic and web dataset with an emphasis on reasoning, making it the most reasoning-focused open-weight small model in the set. The content effect is strongly suppression-dominant (Figure~\ref{fig:delta_grid}), consistent with the model's synthetic training data instilling robust resistance to in-context bias signals, except, again, for Planning Fallacy. The DiD ATT for Planning Fallacy is positive and significant; the SCM gap confirms causality. This robustness coexists with a moderate presence effect, suggesting that the model responds to context presence but filters the bias content effectively for most bias types.

\begin{figure}[h]\centering
\includegraphics[width=0.65\textwidth]{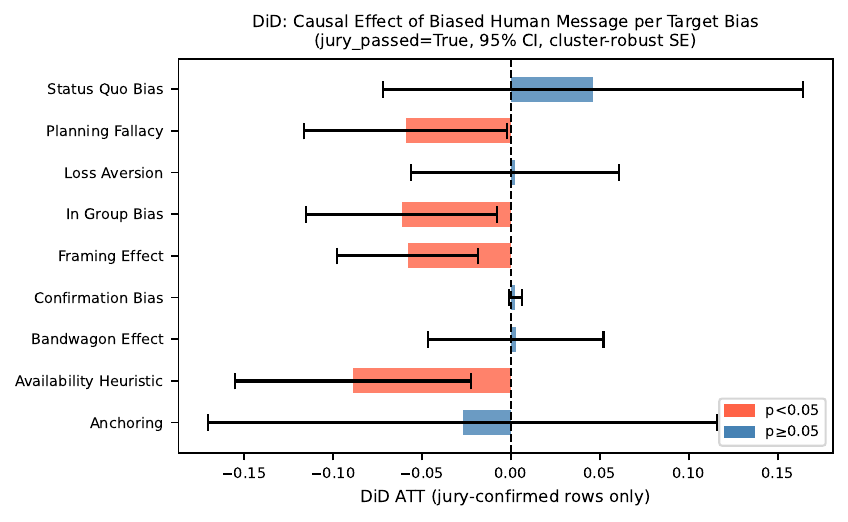}
\caption{Phi-4 DiD ATT. Planning Fallacy is the only significant positive ATT; all other biases show suppression, consistent with the model's reasoning-focused training.}\end{figure}
\begin{figure}[h]\centering
\includegraphics[width=0.65\textwidth]{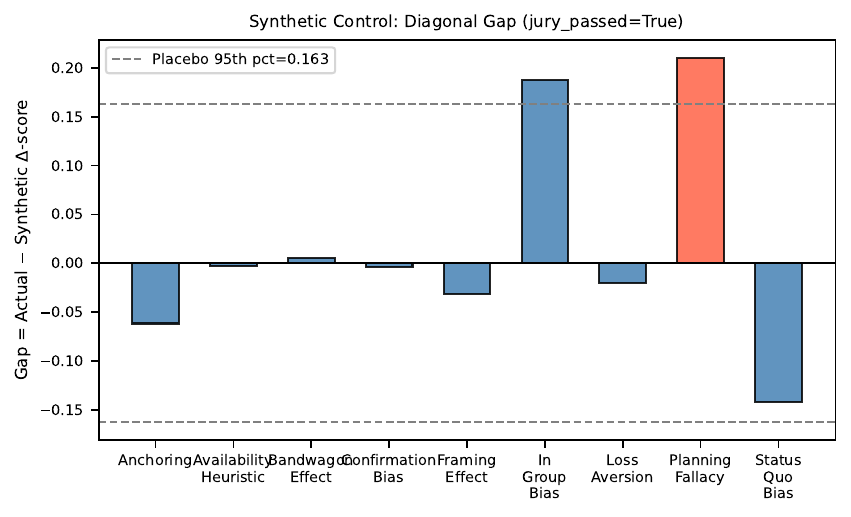}
\caption{Phi-4 SCM gap. Planning Fallacy gap exceeds the placebo distribution; other biases remain within the placebo envelope, confirming selective inducibility.}\end{figure}

\subsection{Gemma-2-9B-IT}

Gemma-2-9B-IT uses a standard open-weight SFT + RLHF recipe and shows a clear suppression-dominant pattern ($\bar{\delta}=-0.028$), with Planning Fallacy the single warm diagonal cell in Figure~\ref{fig:delta_grid}. The DiD and SCM results confirm that the Planning Fallacy amplification is causal and replicable in a 9B-parameter model, ruling out the possibility that the effect is capacity-dependent and visible only in large models.

\begin{figure}[h]\centering
\includegraphics[width=0.65\textwidth]{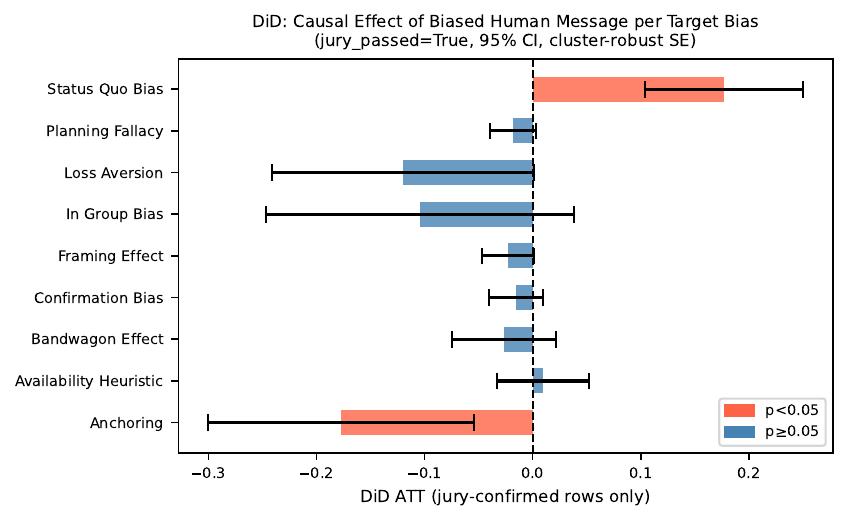}
\caption{Gemma-2-9B-IT DiD ATT. Planning Fallacy is the only significant positive ATT, confirming that the inducible-bias finding is not capacity-dependent.}\end{figure}
\begin{figure}[h]\centering
\includegraphics[width=0.65\textwidth]{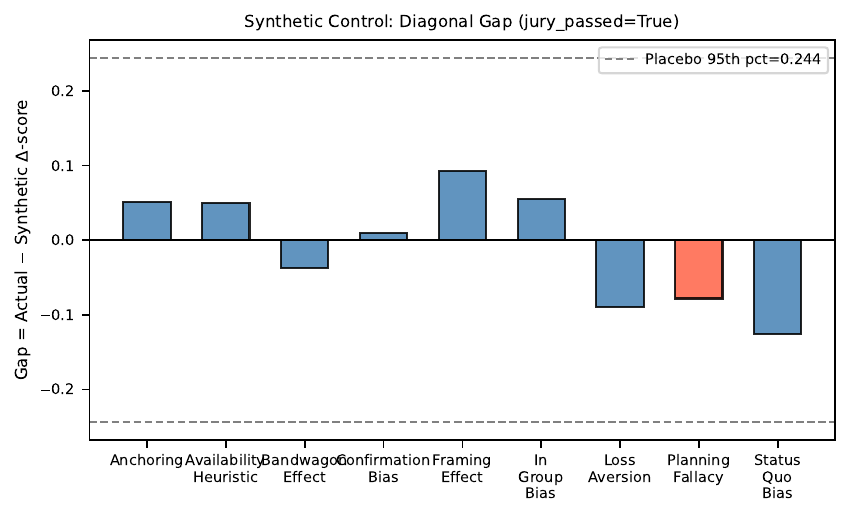}
\caption{Gemma-2-9B-IT SCM gap. Planning Fallacy and Status Quo Bias exceed the placebo distribution; Availability Heuristic and Bandwagon Effect show the most negative gaps, consistent with their suppressible character across all models.}\end{figure}

\end{document}